\documentclass[10pt,twocolumn,letterpaper]{article}

\usepackage[pagenumbers]{cvpr} 

\usepackage{xcolor}
\usepackage{colortbl}

\definecolor{cvprblue}{rgb}{0.21,0.49,0.74}
\usepackage[pagebackref,breaklinks,colorlinks,allcolors=cvprblue]{hyperref}
\usepackage[inline]{enumitem}

\def\paperID{13} 
\def\confName{SLRTP}
\def\confYear{2025}

\title{Exploring Pose-based Sign Language Translation: Ablation Studies and Attention Insights}

\author{Tomáš Železný, Jakub Straka, Václav Javorek, Ondřej Valach, Marek Hrúz, Ivan Gruber\\
Univerzity of West Bohemia,\\ Faculty od Applied Sciences, Department of Cybernetics\\
Univerzitní 2732/8, 301 00 Plzeň, Czech  Republic\\
{\tt\small zeleznyt@kky.zcu.cz, strakajk@kky.zcu.cz, javorek@kky.zcu.cz}\\
{\tt\small valacho@kky.zcu.cz, mhruz@ntis.zcu.cz, grubiv@ntis.zcu.cz}
}

\begin{document}
\maketitle
\begin{abstract}
Sign Language Translation (SLT) has evolved significantly, moving from isolated recognition approaches to complex, continuous gloss-free translation systems. This paper explores the impact of pose-based data preprocessing techniques — normalization, interpolation, and augmentation — on SLT performance. We employ a transformer-based architecture, adapting a modified T5 encoder-decoder model to process pose representations. Through extensive ablation studies on YouTubeASL and How2Sign datasets, we analyze how different preprocessing strategies affect translation accuracy. Our results demonstrate that appropriate normalization, interpolation, and augmentation techniques can significantly improve model robustness and generalization abilities. Additionally, we provide a deep analysis of the model's attentions and reveal interesting behavior suggesting that adding a dedicated register token can improve overall model performance. We publish our code on our GitHub repository\footnote{\url{https://github.com/zeleznyt/T5_for_SLT}}, including the preprocessed YouTubeASL data.
\end{abstract}

\section{Introduction}
Sign language translation has witnessed remarkable progress over the past few decades, transitioning from early isolated sign language recognition systems to more complex continuous recognition frameworks. Early methods largely depended on gloss-based approaches—relying on intermediary linguistic annotations to bridge the visual and textual modalities—while recent research has increasingly shifted toward gloss-free techniques. These gloss-free methods seek to directly map visual inputs into textual outputs, leveraging advances in multi-modal learning and large language models to enhance translation accuracy.

Despite these advancements, gloss-free systems still face notable challenges. Variations in signer position, scale, and background dynamics together with no direct alignment between the input and output languages contribute to the performance gaps when compared to their gloss-based counterparts. In response, our work systematically investigates a series of data preprocessing techniques including keypoint extraction, normalization, and augmentation aiming to mitigate the issues of spatial variation and improve the robustness of the translation pipeline.

We present a comprehensive evaluation of these techniques within a transformer-based framework, specifically adapting a modified T5 encoder-decoder architecture for the task of SLT. Extensive ablation studies are conducted on challenging datasets such as YouTubeASL~\cite{yasl} and How2Sign~\cite{how2sign}, revealing that a thoughtful combination of normalization and augmentation strategies can substantially enhance model performance. Our analysis not only demonstrates improvements in translation accuracy but also provides valuable insights into the interplay between visual preprocessing and model architecture.

Ultimately, this work contributes to the broader goal of developing more accurate and efficient SLT systems, paving the way for enhanced accessibility and communication between the Deaf and hearing communities.

\section{Related Work}
Sign Language Translation has progressed through dynamic evolution over the years, beginning with work on Isolated Sign Language Recognition (ISLR)~\cite{hu2021signbert, li2020transferring} and progressing more towards Continuous Sign Language Recognition (CSLR)~\cite{guo2018hierarchical, camgoz2020sign}, with early efforts primarily focused on isolated sign language (SL) datasets~\cite{desai2023asl, li2020word} and more recent studies advancing with continuous data that capture the dynamic and nature of sign language communication~\cite{PHOENIX-2014T, how2sign, OpenASL, yasl, YouTube-SL-25}. Building on this, SLT has developped in two main approaches: gloss-based and gloss-free methods. Gloss-based approaches utilize structured linguistic representations of signs to learn the alignment between sign language (glosses) and text~\cite{camgoz2020sign, zhou2021improving, chen2022simple, Spoter, zhang2023sltunet}, while gloss-free methods directly map visual features to text, aiming to bypass the need for intermediate linguistic annotations (glosses)~\cite{camgoz2020multi, yin2023gloss, zhou2023gloss, DiffSLT, SignBERT+, guan2024multi}. Gloss-free methods often introduce innovative approaches as for example utilization of self-supervised fine-tuning~\cite{hwang2024universal}, sign pose quantization~\cite{hwang2024gloss} or pseudo-translation tasks~\cite{zheng2023cvt}. Although gloss-based techniques benefit from the transparent supervision, gloss-free approaches have become increasingly popular thanks to advancements in multi-modal learning, with the integration of Large Language Models (LLMs) enhancing the translation accuracy by utilizing better pretrained textual representations~\cite{Sign2GPT, SSVP-SLT, LLaVASLT}.

Thanks to this increasing popularity, we have seen innovations in gloss-free approaches such as Sign2GPT~\cite{Sign2GPT} using large-scale pretrained visual and language models, GFSLT-VLP~\cite{zhou2023gloss} integrating contrastive language–image pretraining and masked self-supervised learning. Innovations went all the way into the topic of diffusion models with DiffSLT~\cite{moon2024diffslt}, a diffusion-based generative approach, transforming random noise into the target latent representation. Furthermore, we also saw SignLLM~\cite{fang2024signllm} applying vector-quantization to convert sign videos into discrete tokens, and SignCL~\cite{ye2024improving} introducing an sign contrastive loss to reduce representation density in dense visual sequences. Moreover, there were innovations such as GASLT~\cite{yin2023gloss}, which incorporates gloss-attention mechanisms, and CSGCR~\cite{zhao2021conditional}, which utilizes custom word verification. Despite these developments and the overall potential, gloss-free SLT methods continue to face a performance gap when compared to their gloss-based alternatives.

Transformer-based models, such as the T5~\cite{T5}, have shown great multilingual capability. Recent literature have explored T5’s flexibility in handling multimodal inputs~\cite{zhang2024scalingsignlanguagetranslation, llmsgoodsignlanguagetranslators, yano2024multilingual}, which showed its potential to address the translation of embedded visual sign language input into text. Additionally, studies using encoder-decoder models that integrate pretrained visual encoders with advanced text decoders — like GFSLT-VLP~\cite{zhou2023gloss} based on mBART~\cite{liu2020multilingual} — indicate that utilizing strong language priors without relying on gloss annotations is an interesting approach to further investigate. Our work uses T5 as model for SL translation and conducts extensive ablation studies which cover areas such as pose augmentation and sign space pose normalization. Recent research shows that while increasing model scale tends to boost performance, using well-curated data and a thoughtfully designed approach is equally important~\cite{khirodkar2024sapiens}.

There are few recent papers related to these topics that explore the utilization of unique pose normalization aiming for encoder-only transformer in SL modeling~\cite{woods2023modelling}, face swapping, and other image (mostly affine) augmentations of SL data which report positive effects during training~\cite{perea2024impact}. Two studies dive into an attention analysis and attention-based sign language recognition built upon decoupled graph and temporal self-attention~\cite{bianco2024signattention, song2022slgtformer}. These studies showcase some interesting observations, for example, that transformer models for SLT learn to attend to sequential clusters rather than individual frames~\cite{bianco2024signattention}, which will be referred to more in Section~\ref{interpretability}.

\section{Methods}
In this section, we describe different parts of our processing parts with emphasis on the parts relevant to the following ablation studies.

\subsection{Data preprocessing}
Data preprocessing is important, especially when working with uncurated datasets. In our experiments, we use YouTubeASL~\cite{yasl} and How2Sign~\cite{how2sign}. YouTubeASL consists of videos captured in the wild and is uncurated, meaning signers appear in various positions, sizes, and resolutions, sometimes alongside other people. In contrast, How2Sign is recorded in a controlled setting with a single signer positioned in front of the camera. However, signers can still shift across videos or appear at different distances.

To address these variations, we first extract keypoints and then evaluate multiple normalization strategies. In both cases, we first split videos into clips based on the captions and work only with the clips.

\subsubsection{Keypoint Extraction} \label{keypoint_extraction}
We use a two-stage approach for keypoint detection: first, we detect a person in the frame, and then we predict keypoints within the detected area. Detecting the person first is crucial, as the signer may occupy only a small portion of the screen (e.g., a news interpreter).

Instead of using a standard object detection model for person detection, we employ a lightweight keypoint detection model. We then define a bounding box around the signer based on the signing space. Signing space is a concept from linguistics, which we define as a rectangle centered between the shoulders, with a width and height four times the shoulder distance. All signing should happen in this area, we make the box slightly bigger than is necessary to ensure that all keypoints are in the box. This guarantees that the signer remains centered, occupies the majority of the frame, and maintains a consistent size across the clip.

We exclude clips containing multiple people, as tracking all individuals across frames and identifying the signer introduces potential errors. To simplify processing, we omit such clips.

Our keypoint extraction pipeline consists of the following steps: 
\begin{enumerate*}
    \item We start by detecting pose using YOLOv8-nano~\cite{Jocher_Ultralytics_YOLO_2023}, if the clip contains multiple people we discard it.
    \item Based on the detected poses we create the signing space.
    \item Next, we spatially crop frames based on the sign space, this ensures that all excessive background is removed and frames are roughly centered on the signer.
    \item Lastly, we use MediaPipe~\cite{lugaresi2019mediapipe} to predict body pose, hand pose, and face mesh in the spatial cropped clip.
\end{enumerate*}

We do not use all keypoints from MediaPipe. For the body pose, we omit leg keypoints, and for the face, we select only a small subset representing prominent facial features. In total, we extract 104 keypoints, this includes 21 keypoints for each hand, 25 for the body pose, and 37 for the face~\footnote{Same as in YouTubeASL paper~\cite{yasl}.}. We use the x and y coordinates generated by MediaPipe, resulting in a final 208-dimensional vector per frame.

\subsubsection{Pose Normalization}
The main step in preprocessing is keypoint normalization, which aims to make keypoints invariant to translation and scale. Although we centered frames on the signer during keypoint extraction, some shifts or size differences may still occur. We evaluate three normalization strategies: two based on the YouTubeASL paper and one based on our signing space approach based on the work~\cite{Spoter}.

In the YouTubeASL paper, normalization is applied by scaling keypoints to fit within a unit bounding box across the entire duration of the clip. We refer to this method as~\textbf{\textit{yasl$_c$}}. This approach ensures that the signer remains of consistent size across all frames but does not account for the changing position within the frame.

\begin{figure}[th]
    \centering 
    \begin{subfigure}[b]{0.44\linewidth}
        \centering
        \includegraphics[width=1\linewidth]{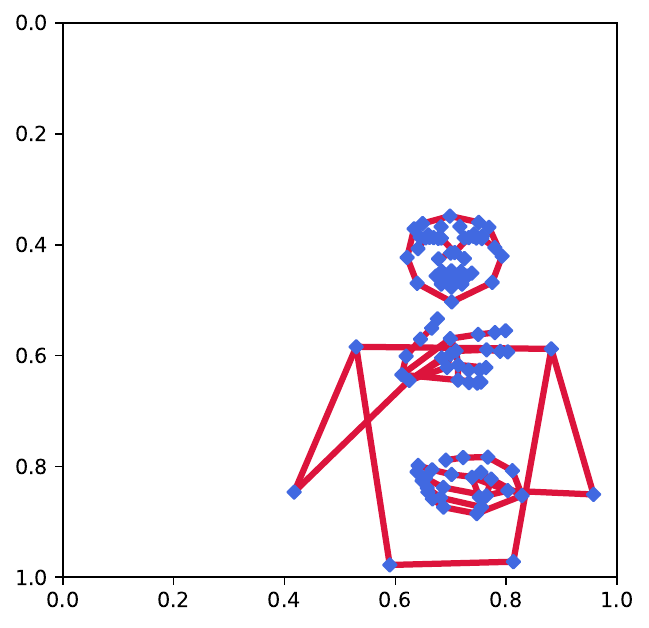}
        \caption{YouTubeASL - Clip\\Normalization}
        \label{fig:yasl_c_norm}
    \end{subfigure}
    \begin{subfigure}[b]{0.44\linewidth}
        \centering
        \includegraphics[width=1\linewidth]{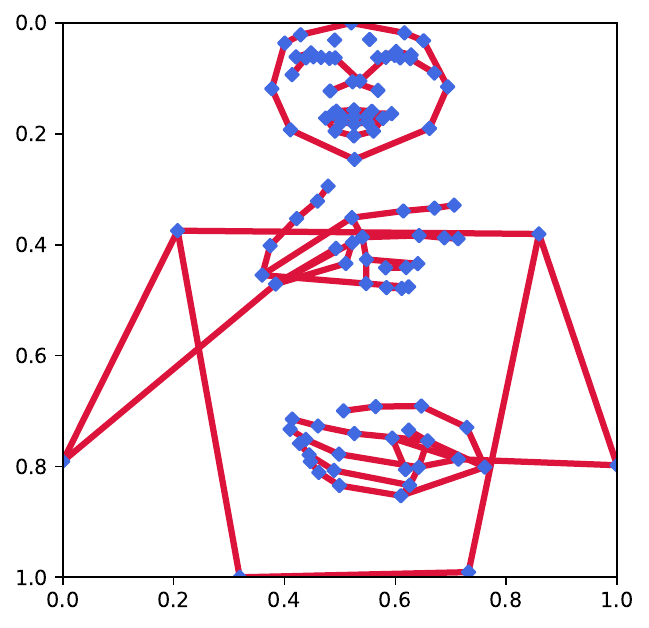}
        \caption{YouTubeASL - Frame\\Normalization}
        \label{fig:yasl_f_norm}
    \end{subfigure}
    \begin{subfigure}[b]{0.9\linewidth}
        \centering
        \includegraphics[width=0.49\linewidth]{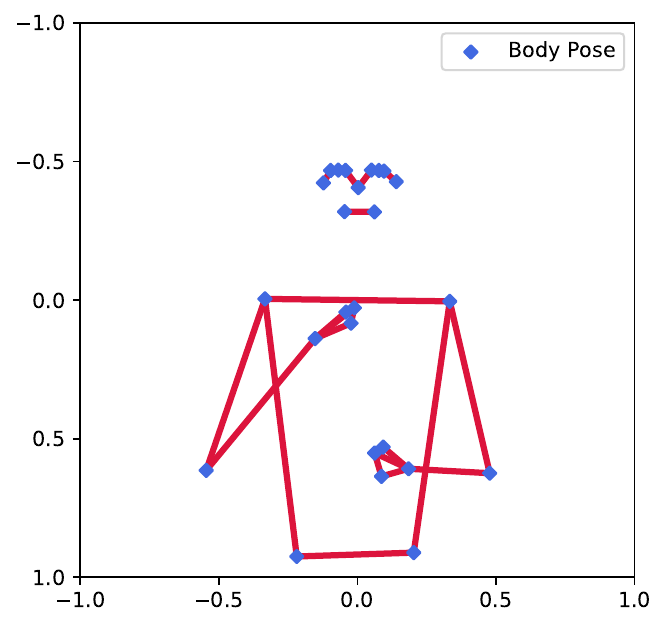}%
        \hfill
        \includegraphics[width=0.49\linewidth]{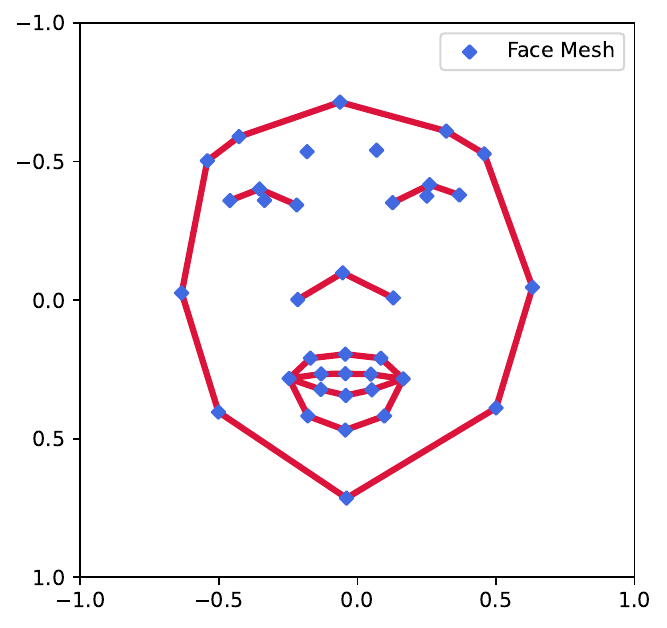}
    \end{subfigure}
    \begin{subfigure}[b]{0.9\linewidth}
        \centering
        \includegraphics[width=0.49\linewidth]{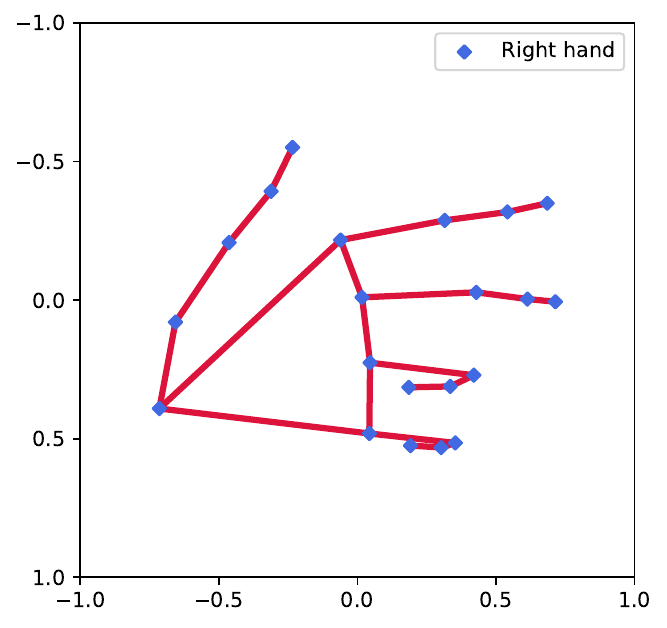}%
        \hfill
        \includegraphics[width=0.49\linewidth]{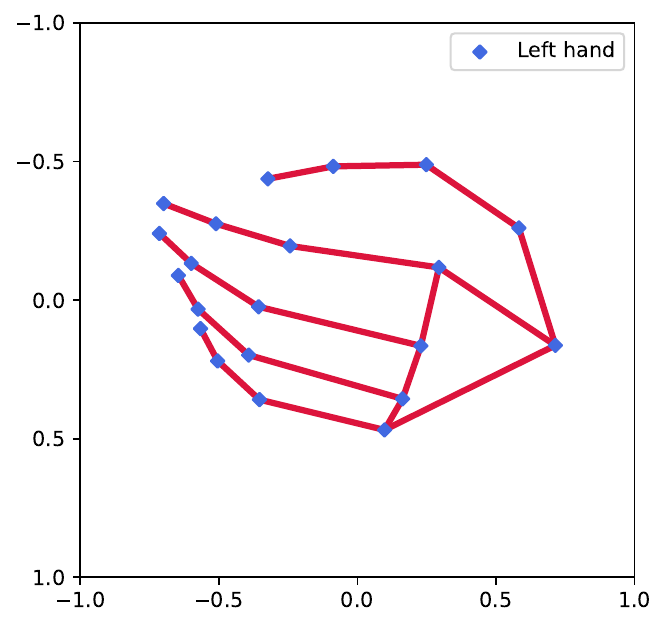}
        \caption{Sign Space Normalization}
        \label{fig:sign_space_norm}
    \end{subfigure}
    \caption{Examples of evaluated normalization methods. We compare multiple approaches: (a) shows the normalization proposed in the YouTubeASL paper, where poses are scaled to fit within a unit box across the entire clip. (b) shows an alternative method where normalization is applied separately to each frame. Finally, (c) illustrates our approach, which normalizes the body pose globally using signing space while applying local normalization separately to the hands and face.}
\end{figure}

We also evaluate a frame-wise normalization strategy, where keypoints are normalized independently in each frame to fit within a unit bounding box. While this method eliminates shifts in the frame and distributes keypoints more evenly within the bounding box, it can cause the signer's size to fluctuate across frames. We refer to this normalization as \textbf{\textit{yasl$_f$}}. Examples of \textit{yasl$_c$} and \textit{yasl$_f$} normalized keypoints are shown in Figure~\ref{fig:yasl_c_norm} and Figure~\ref{fig:yasl_f_norm}, respectively.

The third normalization method (denoted as \textit{SignSpace}) we evaluate is based on the signing space we defined in Subsubsection~\ref{keypoint_extraction}. We normalize body pose keypoints by creating a bounding box centered between the shoulders, with its width and height set to three times the distance between the shoulders. Keypoints within the signing space are then scaled to be in the range $\left\langle -1, 1 \right\rangle$. After scaling, keypoints are shifted so that the center of the signing space is at position $\left[ 0, 0 \right]$. This normalization is applied frame by frame and we consider it as global, as it preserves the relation between the individual body parts.

Global normalization is applied only to body pose keypoints. For hands and face, we use local normalization, meaning we normalize each hand and face separately by scaling them to range $\left\langle -1, 1 \right\rangle$ while maintaining their aspect ratio. Additionally, we add a 10\% border from each side around them to suppress the effect of inaccuracies in the pose estimation model. Local normalization ensures a focused view of individual parts, independent of their absolute position. The absolute position and relationship between different body parts are instead captured through global body pose normalization. Example of keypoints normalized by this method is depicted in Figure~\ref{fig:sign_space_norm}.

\subsubsection{Missing Values}
One important issue that is necessary to handle during the normalization are miss-detections that result in missing values. Some of the keypoints may not be detected, or the signer may use only one hand, with the other hand out of the frame. In the YouTubeASL paper, missing values are handled by replacing them with a large negative value. We adopt this approach, but we also propose to linearly interpolate keypoints if the frame gap between the detected keypoints is short. 

In our case, the gap is two frames or shorter in almost 60\% of cases, and three frames or less in almost 75\% of cases. We assume that the change between close frames is small, which means that interpolated keypoints should be close enough to retain their semantic meaning. We compare this to the approach where all the missing values are replaced with constant values.

\subsubsection{Augmentations}\label{sec:augmentations}
Augmentations are commonly used to enrich datasets. However, in the SLT task, it is essential to ensure that the augmentation process does not change the semantic meaning of the pose. We evaluated three augmentation strategies, each varying the probability and intensity of augmentations by scaling the default strategy values. The default strategy is \textit{heavy}, the second is \textit{medium} (scaled by 0.75), and the last is \textit{light} (scaled by 0.5). 

We use mainly geometric augmentations, which include: rotation, shear, perspective, arm rotation, and additive Gaussian noise. The same augmentations are applied to all frames in the clip. 

\begin{figure}[th]
    \centering 
    \begin{subfigure}[b]{0.44\linewidth}
        \centering
        \includegraphics[width=\textwidth]{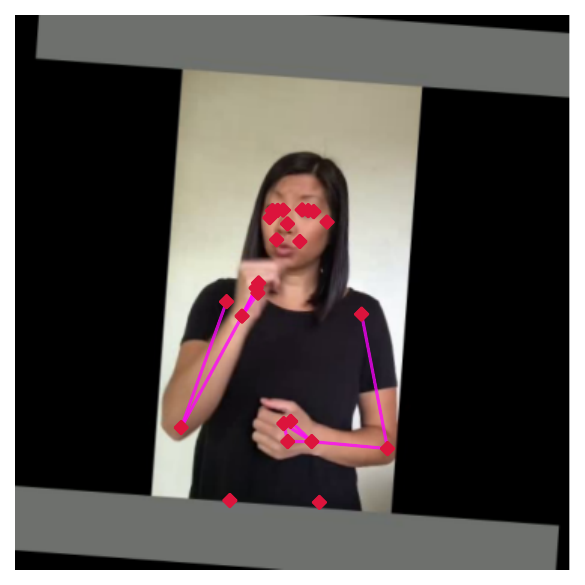}
        \caption{Rotate}
    \end{subfigure}
    \begin{subfigure}[b]{0.44\linewidth}
        \centering
        \includegraphics[width=\textwidth]{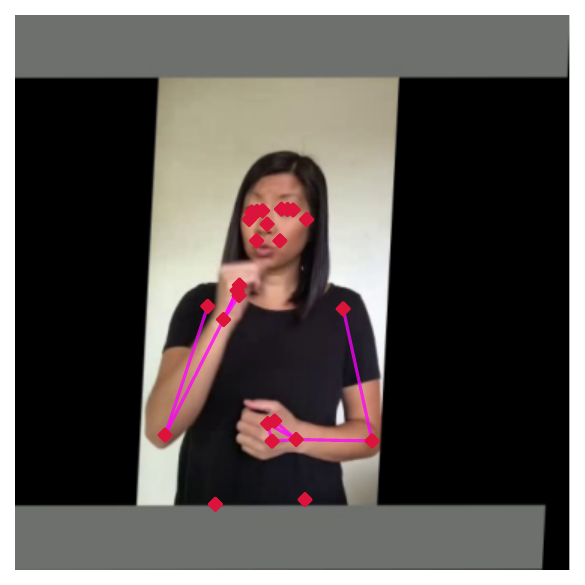}
        \caption{Shear}
    \end{subfigure}
    \begin{subfigure}[b]{0.44\linewidth}
        \centering
        \includegraphics[width=\textwidth]{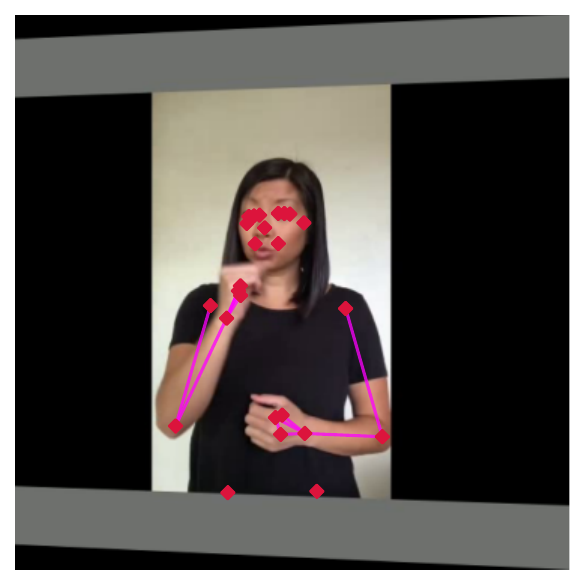}
        \caption{Perspective}
    \end{subfigure}
    \begin{subfigure}[b]{0.44\linewidth}
        \centering
        \includegraphics[width=\textwidth]{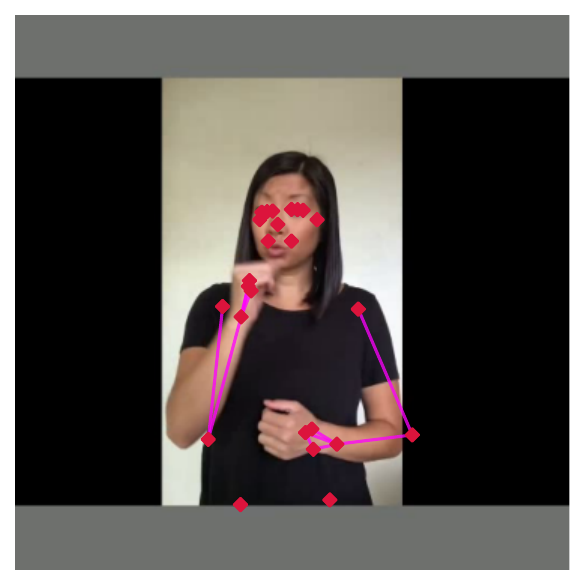}
        \caption{Arm rotation}
    \end{subfigure}
    \caption{Examples of individual augmentations. We show only body pose keypoints, during training all keypoints are augmented. To better illustrate their effects, we applied the same geometric augmentation (except for arm rotation) to the frame.}
    \label{fig:aug}
\end{figure}

Arm rotation augmentation rotates all arm and hand keypoints around a shoulder, elbow, or wrist keypoint. This augmentation can be chained, which means that the entire arm can first rotate around the shoulder, then again around the elbow or wrist in successive transformations. In Figure~\ref{fig:aug}, there are examples of some of the augmentations. Here, each augmentation is applied individually, but during the training, multiple augmentations can be applied to one frame.

\subsection{Model}
Our model setup follows the original baseline method of the YouTubeASL paper. We use a modified version of T5~\cite{T5} encoder-decoder-based transformer. In order to process the input of the 208-dimensional keypoint features, we employ a custom linear layer at the transformer's encoder input instead of traditional tokenized text. Following standard embedding layer practices, our custom layer does not include an additive bias. Besides this change, our model follows a standard T5v1.1\footnote{\url{https://github.com/google-research/text-to-text-transfer-transformer/blob/main/released_checkpoints.md}} architecture. The T5 weights are initialized from T5X, while the custom layer uses the Xavier initialization.

\section{Experiments and Quantitative Results}
In this section, we first describe our experimental setup. Then, we report and analyze the results of three different ablation studies.

\subsection{Experimental Setup}
In our experiments, we finetune the T5-based model on the YouTubeASL dataset using various data preprocessing techniques to evaluate their impact on overall model performance. For YouTubeASL we use a custom 90:10 train-val split, while for the How2Sign dataset we use the default split provided by the dataset. All our experiments are assessed based on the BLEU scores computed using sacrebleu v2.4.3 on the How2Sign dataset, a standard benchmark for gloss-free sign language translation systems. The performance on the How2Sign dataset is measured without any additional finetuning on this dataset. If not stated otherwise, the model is trained for a total of 200,000 iterations using an effective batch size of 256 and a constant learning rate of 0.0004. In the initial experiments, we observe high training volatility. To reduce this variability between training runs, we employ a warm-up phase for the first 5,000 training steps. Additionally, to ensure a fair comparison between different training setups, we run each experiment with three different seeds and report the best run. The YouTubeASL paper doesn't provide the exact value to use in case of missing keypoint values. Inspired by their mention of a "large negative number," we use a value \texttt{-10} in our experiments. The training was conducted using 4 AMD MI250x GPU modules, split into 8 GCD for each experiment.

It should be noted that the trained models after 200,000 iterations are not "fully trained", and their performance would benefit from additional training; there are two reasons for this shorter training protocol. Firstly and more importantly, we believe the comparative performance after this shorter training protocol reflects the performance comparison of fully-trained models. The second reason is based on the restriction of computational resources available. 

\subsection{Normalization}
In the first set of experiments, we analyze four different types of data normalization. Results can be seen in Table~\ref{tab:normalization}. 


\begin{table}[]
\centering
\begin{tabular}{l|ccccc}
\toprule
Normalization& B-1      & B-2 & B-3 & B-4 \\ \hline  \hline
none    & 13.62   & 3.67   & 1.54   & 0.73 \\  \hline
\textit{yasl$_c$}    & 13.00   & 3.90   & 1.59   & 0.66 \\
\textit{yasl$_f$}    & 14.67   & 4.78   & 2.19   & 1.13 \\
\textit{SignSpace}    & \textbf{17.47}   & \textbf{7.19}   & \textbf{3.79}   &\textbf{ 2.17} \\ \hline
\end{tabular}
\caption{Comparison of four different types of normalization techniques. Performance is measured by BLEU scores on the How2Sign dataset.}
\label{tab:normalization}
\end{table}

All the proposed normalization results in better performance when compared to the training without any normalization. Interestingly, the original \textit{yasl$_c$} performs worse than our modification \textit{yasl$_f$}. We argue that the speaker size change in the \textit{yasl$_f$} normalization is less \textit{distracting} for the model than the shift in the speaker position in \textit{yasl$_c$}. The \textit{SignSpace} normalization outperforms all other normalization approaches by a large margin. Based on this result, all the following experiments use the \textit{SignSpace} normalization.

\subsection{Interpolation}
In the next series of experiments, we analyze the effect of using linear interpolation of the missing keypoints. We experiment with a total of 3 different settings: interpolate all gaps with size 2 or smaller, with gaps 3 or smaller, or don't use interpolation at all, in which case all missing values are replaced with the default value equal to $-10$. The results are in Table~\ref{tab:interpolation}.


\begin{table}[]
\centering
\begin{tabular}{l|ccccc}
\toprule
Interpolation& B-1      & B-2 & B-3 & B-4 \\ \hline  \hline
none    & \textbf{17.47}   & 7.19   & 3.79   & 2.17 \\  \hline
$\leq$2 frames    & 16.91   & 7.35   & \textbf{4.06}   & \textbf{2.43} \\
$\leq$3 frames    & 17.16   & \textbf{7.40}   & 4.01   & 2.33 \\ \hline
\end{tabular}
\caption{Comparison of three different interpolation settings. Performance is measured by BLEU scores on the How2Sign dataset.}
\label{tab:interpolation}
\end{table}

Both interpolation approaches result in slightly better results than runs without any interpolation. We hypothesize that the interpolation makes data easier to interpret and additionally gives the model more frames where relevant information is stored.

\subsection{Augmentations}
We investigate the model's performance using different types of augmentations. First, we assess the contributions of individual augmentations by applying them with a medium-scale value and evaluating the finetuned models. Based on these individual performances, we select those augmentations that positively impact performance to design a final augmentation protocol with three different scales, as described in Section \ref{sec:augmentations}. 

According to Table~\ref{tab:augmentaions_individual}, 
the overall performance (majority of the BLEU scores) was improved by the shear, rotate elbow, and noise augmentations. In our final augmentation protocols, we used only these three types of augmentation. We tried to analyze the other augmentations and their effect on the inputs. The decrease in performance for the rotate augmentation is probably caused by the fact that rotation is not very common in real-world data examples. Therefore, it does not contribute to the necessary generalization and only makes the training data more difficult. The same is true for the perspective augmentation. Additionally, we argue that augmentation of the shoulder and wrist rotation can be too heavy in the sense that they can easily change the meaning of signs. 

\begin{table}[]
\centering
\begin{tabular}{l|ccccc}
\toprule
Augmentation & B-1  & B-2 & B-3 & B-4 \\ \hline  \hline
none    & 17.47   & 7.19   & 3.79   & 2.17 \\\hline
rotate    & \textcolor{red}{15.30}   & \textcolor{red}{5.73}   &\textcolor{red}{2.88}   & \textcolor{red}{1.61} \\
shear    & \textcolor{red}{17.19}   & \textcolor{ForestGreen}{7.25}   & \textcolor{ForestGreen}{3.86}   & \textcolor{ForestGreen}{2.2} \\
perspective    & \textcolor{red}{16.07}   & \textcolor{red}{6.83}   & \textcolor{red}{3.70}   & 2.17 \\
rotate shoulder    & \textcolor{red}{16.39}   & \textcolor{red}{6.97}   & \textcolor{red}{3.75}   & 2.17 \\
rotate elbow    & \textcolor{ForestGreen}{17.48}   & \textcolor{ForestGreen}{7.38}   & \textcolor{ForestGreen}{3.89}   & \textcolor{ForestGreen}{2.28} \\
rotate wrist    & \textcolor{red}{16.05}   & \textcolor{red}{6.84}   & \textcolor{red}{3.72}   & \textcolor{ForestGreen}{2.20} \\
noise    & \textcolor{red}{17.45}   & \textcolor{ForestGreen}{7.47}   & \textcolor{ForestGreen}{4.07}   & \textcolor{ForestGreen}{2.41} \\
\hline
\end{tabular}
\caption{Impact of individual augmentations. Performance is measured by BLEU scores on the How2Sign dataset.}
\label{tab:augmentaions_individual}
\end{table}

The final results of our three augmentation protocols are presented in Table~\ref{tab:augmentations_protocols}.



\begin{table}[]
\centering
\begin{tabular}{l|ccccc}
\toprule
Augmentations& B-1      & B-2 & B-3 & B-4 \\ \hline  \hline
none    & \textbf{17.47}   & 7.19   & 3.79   & 2.17 \\\hline
light  &  15.76  &  6.23  &  3.12  &  1.71 \\
medium  &  17.27  &  \textbf{7.51}  &  \textbf{4.12}  &  \textbf{2.46} \\
heavy  &  16.58  &  7.10  &  3.85  &  2.29 \\ \hline
\end{tabular}
\caption{Impact of different augmentation protocols. Performance is measured by BLEU scores on the How2Sign dataset.}
\label{tab:augmentations_protocols}
\end{table}

Based on the results, it seems that the medium augmentation protocol slightly improves the final results. The other two protocols are comparable with the setup without any augmentations. There are two main possible reasons why this phenomenon occurred. First, our training protocol is too short. Based on the analysis of training curves, we do not see any saturation in the results. The lack of saturation, in conjunction with the fact that training with augmentations is generally slower due to the increased complexity of the training set, could result in worse performance after a certain number of iterations. Second, the YouTubeASL dataset is a very complex dataset with a large number of data samples. Therefore, the proposed augmentation may not bring any helpful information into the training. We want to analyze this phenomenon more in our future research.

\section{Qualitative Results}
In this section we provide qualitative results in form of self- and cross-attention analysis of our T5v1.1-base model. We also analyze translations that are learned on the weakly aligned data from the YouTubeASL dataset. 

\subsection{Encoder Self-Attention}\label{self-attn}
\begin{figure}[th]
    \centering
    \includegraphics[width=1\linewidth]{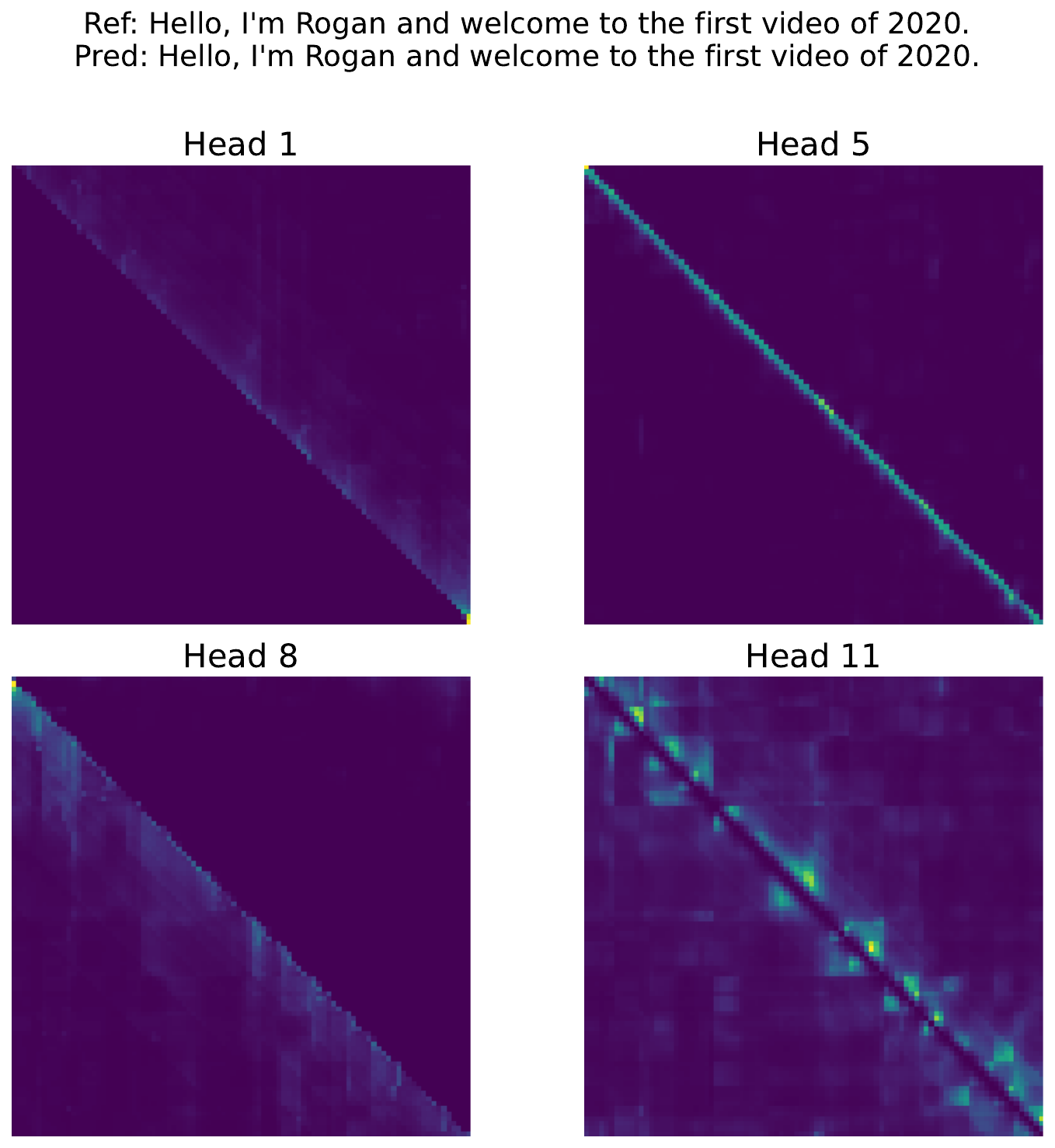}
    \caption{Encoder self-attention averaged over layers per attention head. We observe that while Head 5 strictly focuses on the current token (visible as attention along the diagonal), Heads 1 and 8 specialize in attending to past and future contexts, respectively. Head 11, on the other hand, exhibits a more complex pattern, attending broadly to the surrounding context beyond the immediate diagonal.}
    \label{fig:enc-attn}
\end{figure}

To analyze the patterns in the encoder attention mechanism during T5 inference, a visualization averaged over all encoder layers (Figure~\ref{fig:enc-attn}) shows that each of the 12 attention heads specializes in identifying a distinct causal pattern within the input signal. Furthermore, each head focuses on a different temporal context surrounding the current frame. These findings stand true for all analyzed data hinting at a learned specialization of each head. More examples with all attention heads visualized can be found in the supplementary material.

\subsection{Cross-Attention Behavior}\label{cross-attn}
In the cross-attention matrices during inference, we demonstrate a clear causal relationship between encoder and decoder representations. The attention progresses sequentially over time, consistent with the linear advancement of both textual and ASL signals, resulting in an attention distribution that disperses over segmented words, as we present in a selected cross-attention matrix in Figure~\ref{fig:cross-attn_layer}. The other layers' visualization can be found in supplementary material.

\begin{figure}[th]
    \centering
    \includegraphics[width=1\linewidth]{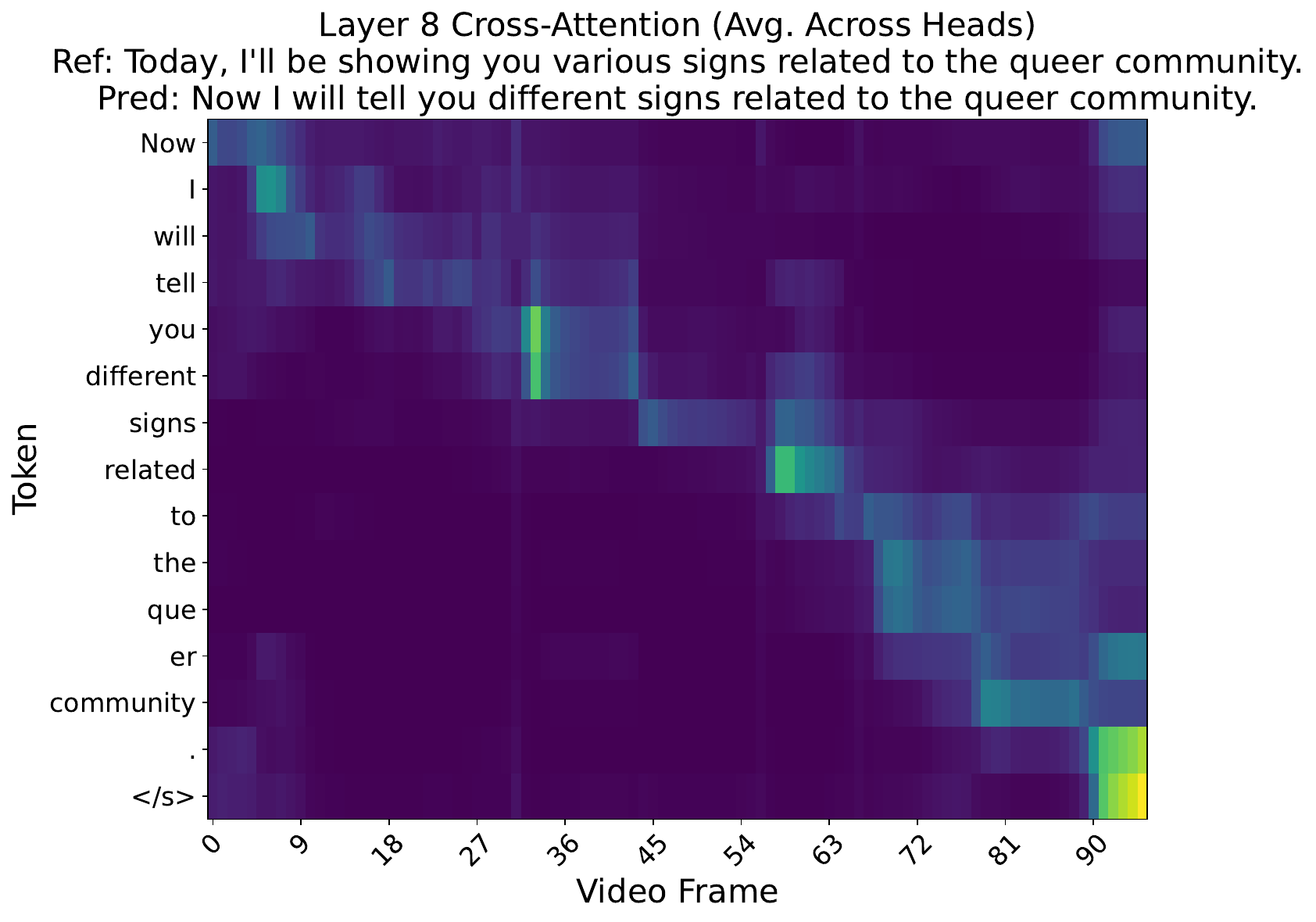}
    \caption{Cross-Attention averaged over all attention heads in a layer, showing temporal progression of tokens attending to frames.}
    \label{fig:cross-attn_layer}
\end{figure}

\begin{figure}[th]
    \centering
    \begin{subfigure}{\linewidth}
        \centering
        \includegraphics[width=\linewidth]{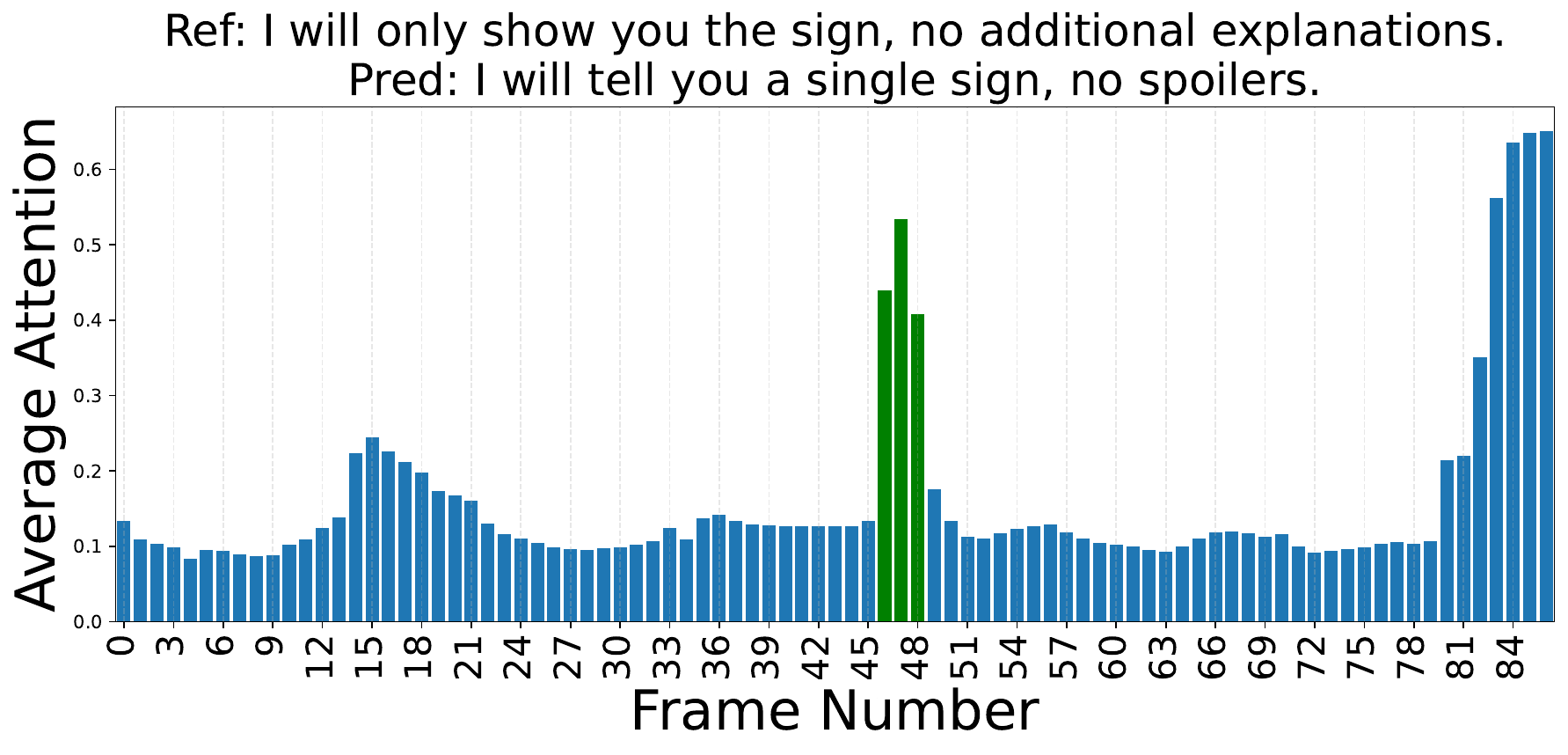}
        \caption{}
        \label{fig:cross-attn_spike_histogram}
    \end{subfigure}
    
    \begin{subfigure}{\linewidth}
        \centering
        \includegraphics[width=\linewidth]{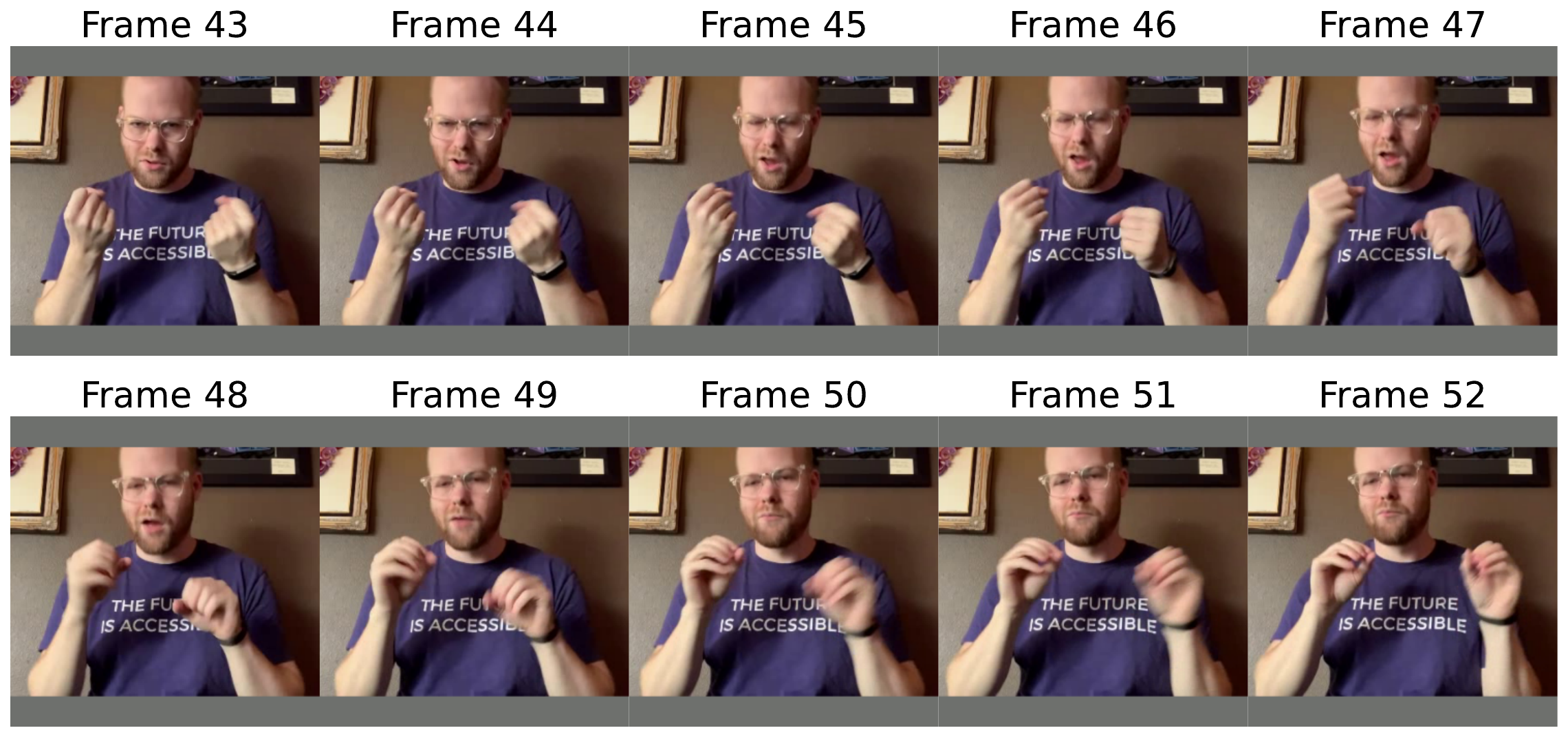}
        \caption{}
        \label{fig:cross-attn_spike_frames}
    \end{subfigure}
    
    \caption{Histogram (a) visualizes Cross-Attention Distribution over all attention heads and layers, with an intensity spike in frames 46–48, highlighted in green. In (b), the corresponding video frames show the keyframe for the word "SIGN" matches the time of the cross-attention spike.}
    \label{fig:cross-attn_spike}
\end{figure}

Next to this, we have revealed another kind of trend in the cross-attention data. In majority of the analyzed matrices, averaged across heads and layers, there appears to be a spike in intensity in the last few frames towards the end of the clip. Also, in many clips there is an attention spike in several other places across the clip. This behavior suggests that the decoder is placing greater attention on a specific subset of input frames when generating each decoded token. This can be observed in Figure~\ref{fig:cross-attn_spike_histogram}. When the spike appears during the signing we found out that it is usually located around a key-sign of the utterance where no transition between signs occurs. This is an expected behavior in the task of SL translation. However, this does not explain the consistent behavior of the high peaks at the end of the utterance observed in almost every clip. 

Upon further examination, we found clips that had high cross-attendances to long segments in various parts of the input, Figure~\ref{fig:cross-attn_sequence_histogram} and more in supplementary. When we investigated these clips, we were surprised that the decoder was attending the part of the clip where no signing was performed. This led us to a hypothesis that the T5 model is using these non-informative segments to encode crucial information about the translation. This behavior has been already observed in previous works~\cite{bondarenko2023quantizable, darcet2024visiontransformersneedregisters} where they use register buffers as additional tokens to encode such information. In the work~\cite{darcet2024visiontransformersneedregisters} the analysis is performed over images where the model is usually encoding important information in patches belonging to the background. This would be analogous to our observations and it might be helpful to use the same principle of adding register buffers to our translation model for better interpretability and generalization. 



\begin{figure}[t!]
    \centering
    \begin{subfigure}{\linewidth}
        \centering
        \includegraphics[width=\linewidth]{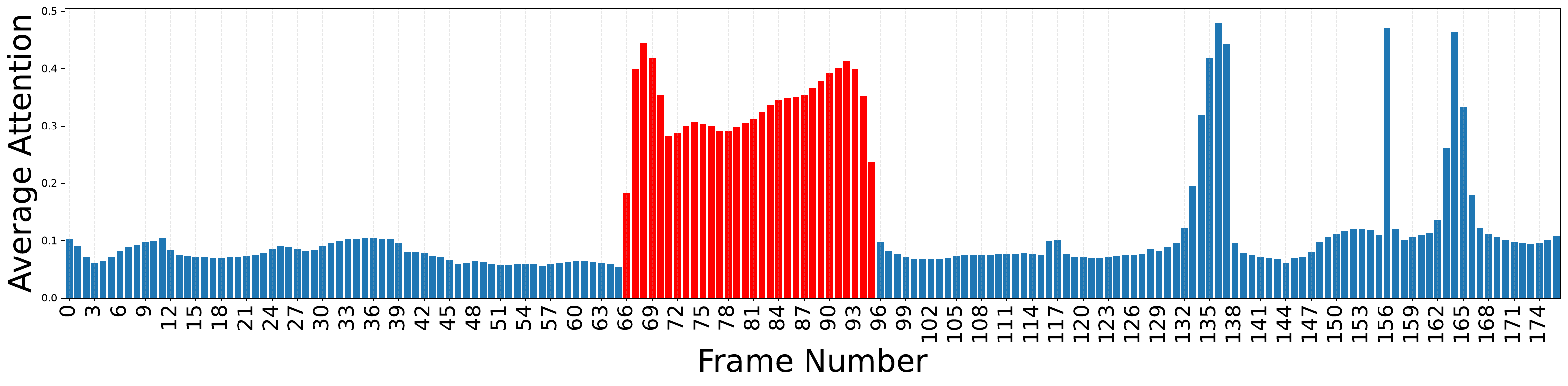}
        \caption{}
        \label{fig:cross-attn_sequence_histogram}
    \end{subfigure}
    
    \begin{subfigure}{\linewidth}
        \centering
        \includegraphics[width=\linewidth]{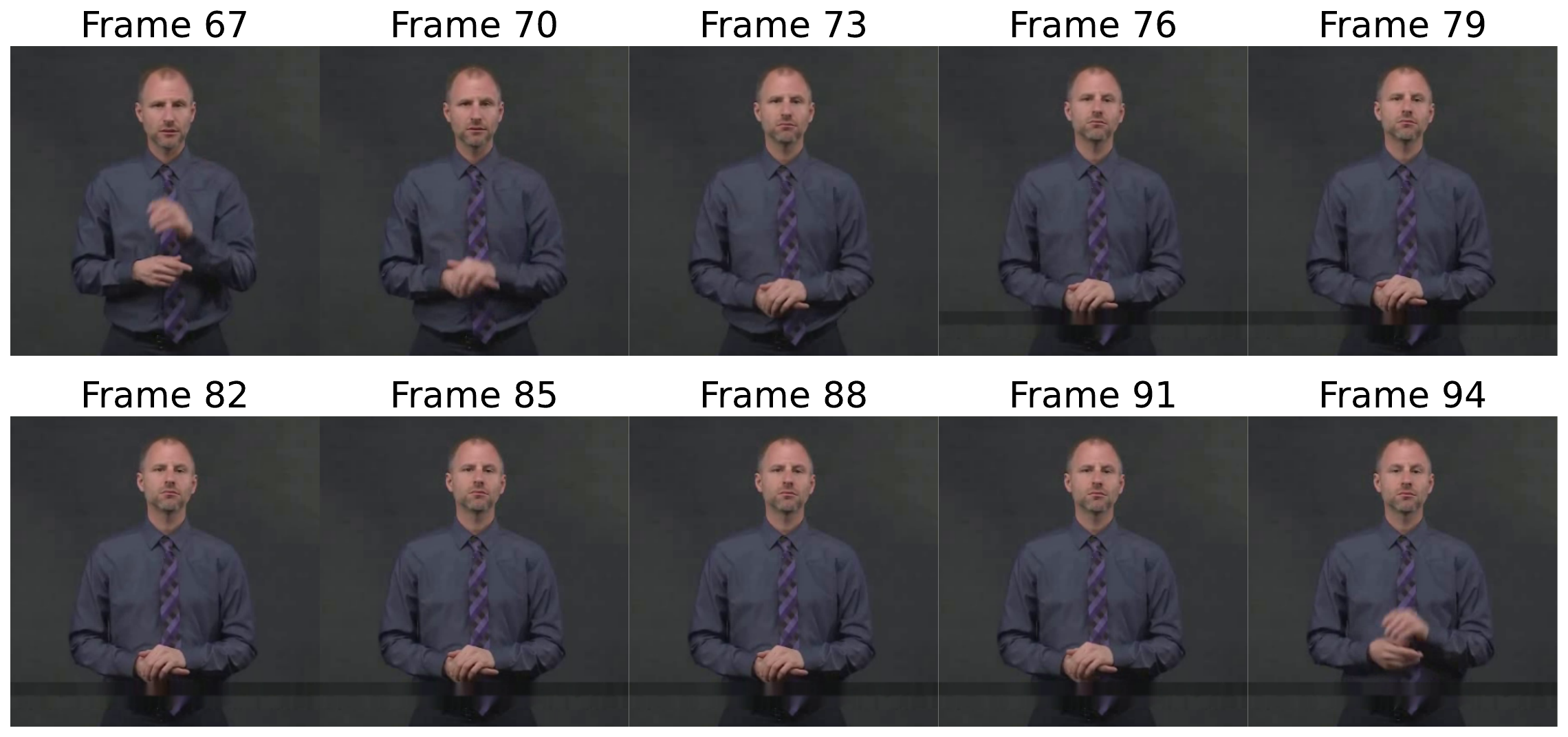}
        \caption{}
        \label{fig:cross-attn_sequence_frames}
    \end{subfigure}
    
    \caption{Histogram (a) visualizes Cross-Attention Distribution over all attention heads and layers, with a long intensity spike sequence in frames 66-95, highlighted in red. Video frames (b) show this is a sequence of still, non-informative frames of a transition.}
    \label{fig:cross-attn_sequence}
\end{figure}

\subsection{Integrated Gradients Analysis}\label{interpretability}

Another standard approach to analyzing the model's behaviors is an analysis of integrated gradients. In this paper, we utilized Captum library~\cite{kokhlikyan2020captum} to perform gradient analysis and assign attribution scores to input features. To be more specific, we used the Integrated Gradients tool, which accumulates gradients along a linear path from a baseline (in our case, an array of zeros) to the actual input, assigning an attribution score to each frame for the final prediction. These scores reveal which frames positively or negatively influence the model's translations, among other things also supporting our observations in Sections~\ref{self-attn} and \ref{cross-attn}. We used well-translated test samples only to clearly correlate positive attributions with high-quality translations. Positive attributions, therefore, indicate that certain frames aid in accurate translations, while negative scores may reflect noise or temporal misalignment; examples are shown below.

\begin{figure}[th]
    \centering 
    \includegraphics[width=1\linewidth]{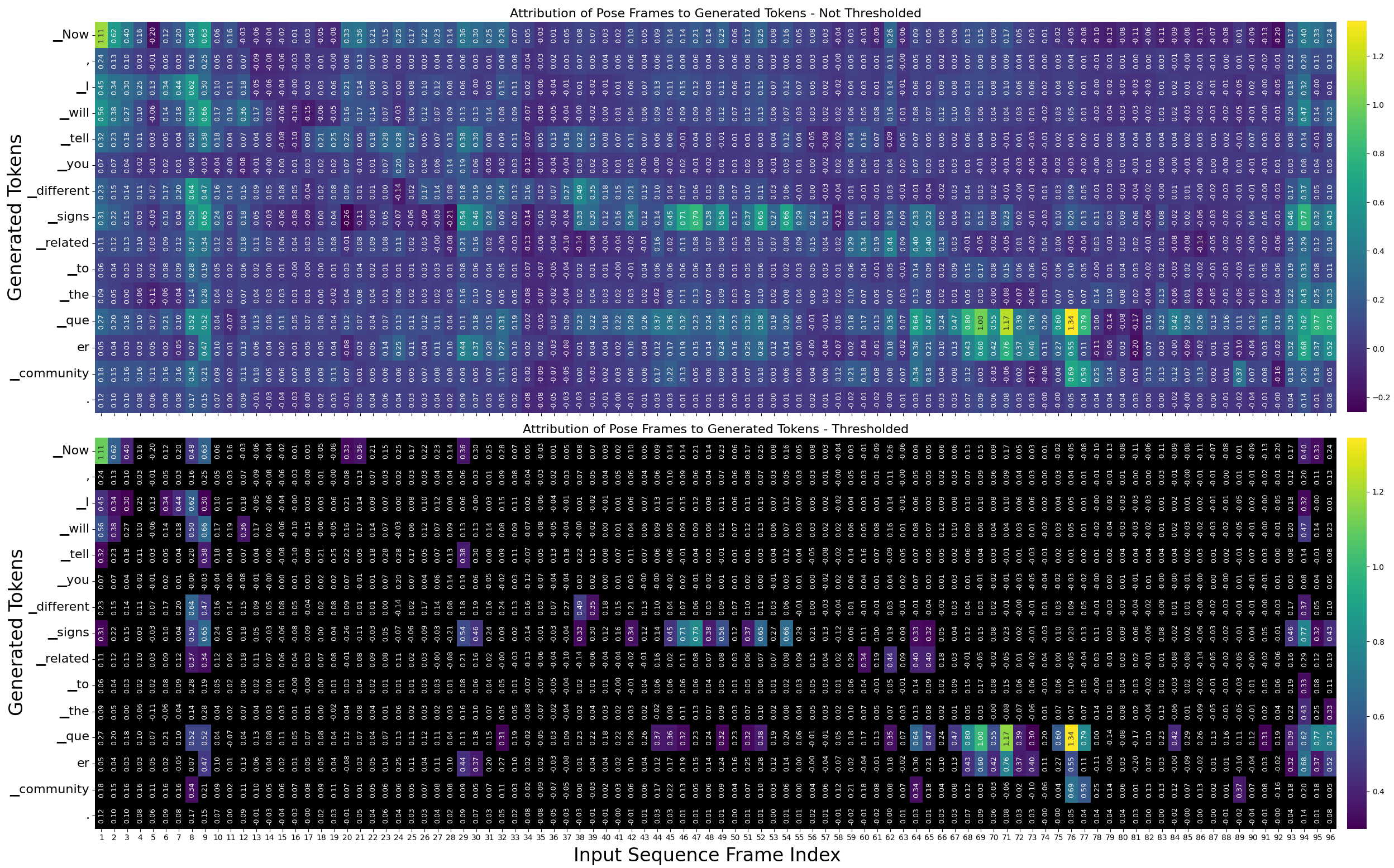}
    \caption{Attribution of Pose Frames to Generated Tokens (top): our finetuned T5 model for SLT translating a chosen phrase, (bottom): the identic model and phrase with minimal threshold of 0.3 to better showcase the diagonal trend.}\label{interpret}
\end{figure}

\begin{itemize}
    \item \textbf{Reference: } "Today, I'll be showing you various signs related to the queer community."
    \item \textbf{Prediction: } "Now, I will tell you different signs related to the queer community."
\end{itemize}

and the integrated gradients per output token per input frame are shown in Figure~\ref{interpret}.

We observe behavior that is challenging to fully analyze, yet it is noteworthy that it has not been observed for the base (non-finetuned) model. A diagonal trend in integrated gradients is starting to occur. We set an experimental minimal threshold of 0.3 for visualization, see lower Figure in~\ref{interpret}. Two clusters emerge for the tokens “signs” (around index 47) and “que” (around index 70). Punctuation marks (dots and commas) show near-zero contributions, suggesting that while the model retains T5’s textual and textual structure understanding, these punctuation marks are not semantically encoded in the input frames. This indicates that frame importance aligns with the temporal occurrence of signing, whereas off-diagonal patches may reflect contextual influences or incomplete model adaptation.

\begin{figure}[th]
    \centering 
    \includegraphics[width=1\linewidth]{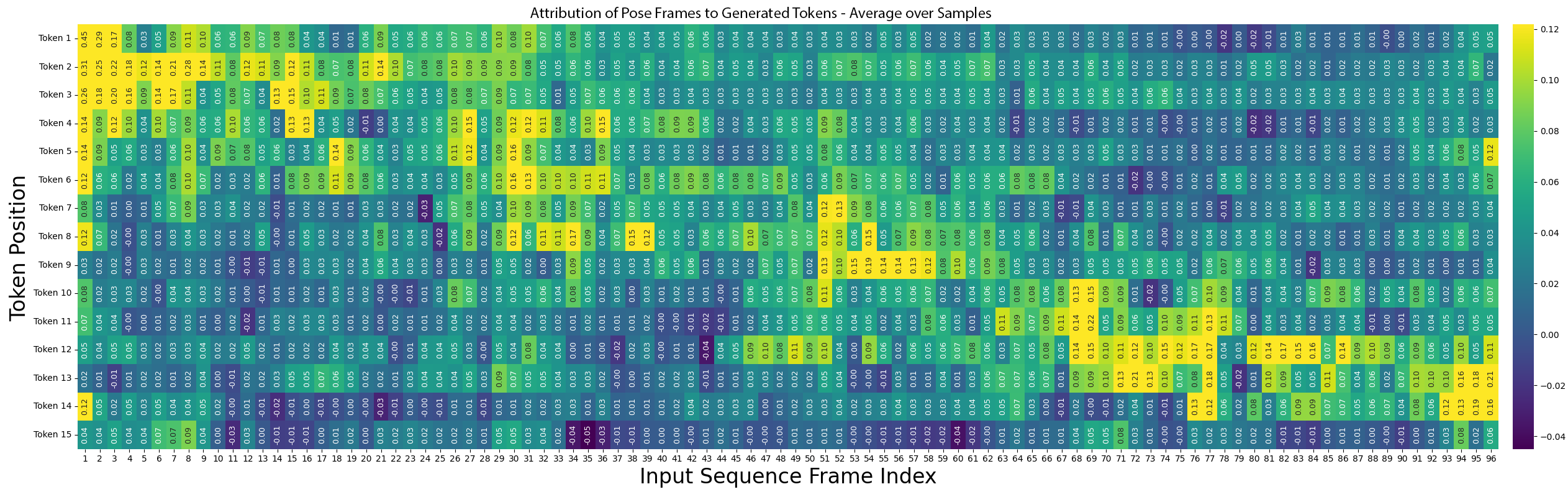}
    \caption{Attribution of Pose Frames to Generated Tokens - filtered average over multiple data samples}\label{IG-average}
\end{figure}

As a following step in the gradients analysis, we performed integrated gradients average over multiple relatively better translated data samples. These were chosen with the rule of a minimal BLEU-1 translation score of 10. In Figure~\ref{IG-average} we observe not just a clear diagonal trend with some integrated gradient clustering tendencies (that aligns with the observation from a SignAttention study~\cite{bianco2024signattention}), but also attributions of multiple last frames to some of the predicted token positions as already discussed in Section~\ref{cross-attn} and also seen from a single data sample analysis, Figure~\ref{interpret}.

\subsection{Analysis of Generalization Capabilities}
\begin{figure}[th]
    \centering
    \begin{subfigure}{\linewidth}
        \centering
        \includegraphics[width=\linewidth]{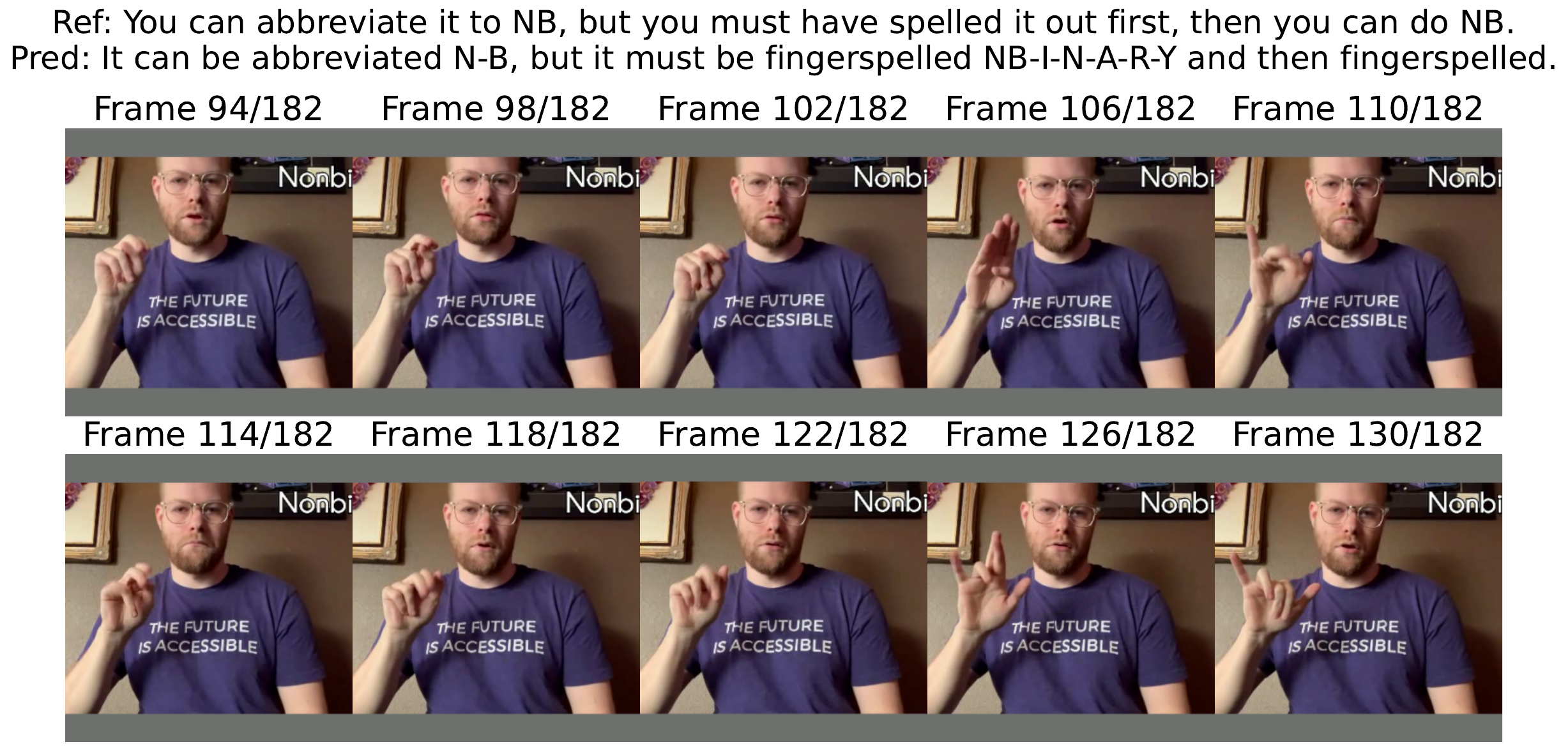}
        \caption{}
        \label{fig:fingerspelling}
    \end{subfigure}
    
    \begin{subfigure}{\linewidth}
        \centering
        \includegraphics[width=\linewidth]{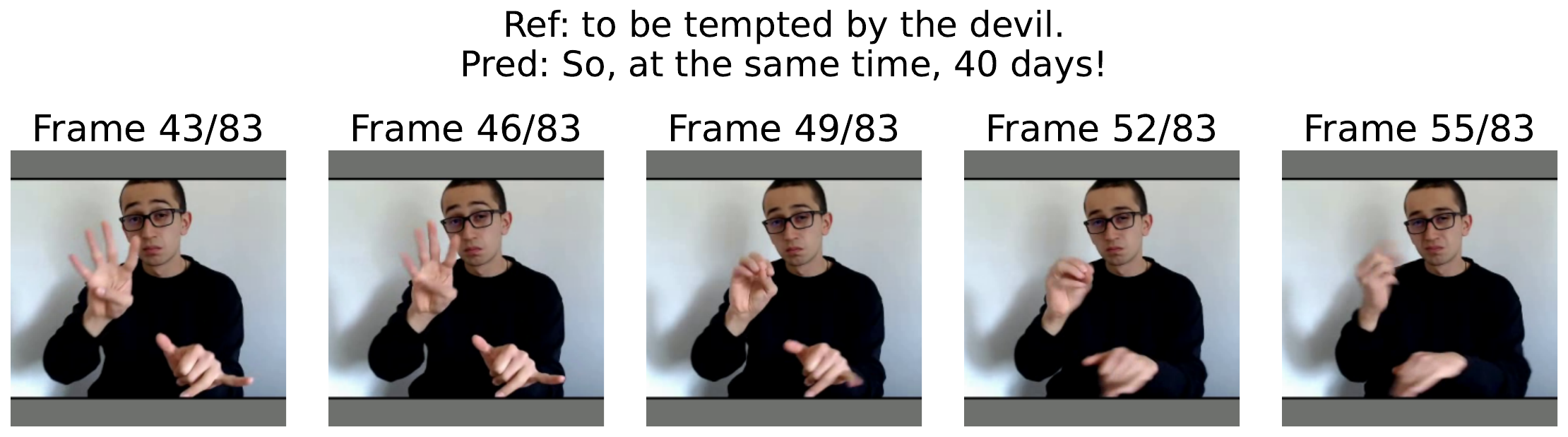}
        \caption{}
        \label{fig:numerals}
    \end{subfigure}
    
    \caption{Example video frame sequences where the model has overcome wrong labels and correctly recognized (a) fingerspelling and (b) numerals.}
    \label{fig:generalization}
\end{figure}

In some cases we have observed that the predicted translations have surprisingly surpassed the reference ones. As YouTubeASL is a weakly-aligned dataset, not all translation labels (taken from video captions) are always correct. For example, the model correctly recognized and translated fingerspelling (Figure~\ref{fig:fingerspelling}) and the signs for numerals (Figure~\ref{fig:numerals}), which were labeled incorrectly and not even present in the reference translation. The reference pushes the gradients in a wrong direction while the model is being optimized. It might be helpful to automatically re-label some dataset samples using machine translated pseudo-labels. Similar ideas were presented in many fields, for SLT notably in~\cite{zhou2019dynamic}. A mechanism that would be able to detect relevant samples and decide which pseudo-labels to use would need to be implemented and will be the subject of our future work.

\section{Conclusion}
This study systematically explored the impact of pose-based preprocessing techniques on Sign Language Translation while using a T5-based model. In extensive ablation studies, we demonstrated the importance of normalization, interpolation, and augmentation techniques. These techniques can significantly impact model robustness, mitigating signer variability and spatial inconsistencies. The ablation studies highlight the effectiveness of normalization based on signing space, interpolation of missing keypoints, and suitable augmentation protocol. Moreover, attention analysis revealed valuable insights into model behavior, suggesting that register tokens could further enhance SLT performance. 

In our future work, we would like to focus on incorporating register tokens and evaluating their influence on SLT accuracy. Furthermore, we would like to explore the possibility of using appearance-based features, such as MAE or DINO features, as additional input into the model.

\section*{Acknowledgment}
The work has been supported by the grant of the University of West Bohemia, project No. SGS-2025-011. Computational resources were provided by the e-INFRA CZ project (ID:90254), supported by the Ministry of Education, Youth and Sports of the Czech Republic.


{
    \small
    \bibliographystyle{ieeenat_fullname}
    \bibliography{main}
}


\end{document}



\section*{Supplementary Material}

\section{Augmentations}
The detailed augmentation protocols are presented in Table \ref{tab:sup_augmentations}. We use standard geometric augmentations. The rotate augmentation rotates all keypoints around the center of the bounding box derived from the body pose keypoints. Shear is applied along either the x- or y-axis. Perspective transformation is applied to either the top and bottom or the left and right sides. Selected side is randomly reduced by a portion from the interval. Arm rotation is applied independently to the shoulder, elbow, and wrist. Finally, noise is added to all keypoints individually.

\begin{table*}[h!]
    \centering
    \renewcommand{\arraystretch}{1.2}
    \begin{tabular}{l l | c c c c}
        \toprule
        \textbf{augmentation} & \textbf{parameter} & \textbf{heavy} & \textbf{medium} & \textbf{light} \\
        \midrule
        rotate & angle & $(-6,6)$ & $(-4.5,4.5)$ & $(-3,3)$ \\
               & prob. & $1.0$    & $0.75$    & $0.50$ \\
        \midrule
        shear & angle x & $(-6,6)$ & $(-4.5,4.5)$ & $(-3,3)$ \\
              & angle y & $(-6,6)$ & $(-4.5,4.5)$ & $(-3,3)$ \\
              & prob. & $0.75$    & $0.56$    & $0.38$ \\
        \midrule
        perspective & portion & $(-0.15,0.15)$ & $(-0.11,0.11)$ & $(-0.08,0.08)$ \\
                    & prob. & $0.50$    & $0.38$    & $0.25$ \\
        \midrule
        rotate arm & shoulder  & $(-10,10)$ & $(-7.5,7.5)$ & $(-5,5)$ \\
                    & elbow  & $(-10,10)$ & $(-7.5,7.5)$ & $(-5,5)$ \\
                    & wrist & $(-10,10)$ & $(-7.5,7.5)$ & $(-5,5)$ \\
                    & prob. & $0.75$    & $0.56$    & $0.38$ \\
        \midrule
        noise & standard dev. & $1.5$ & $1.5$ & $1.5$ \\
              & prob. & $0.75$    & $0.56$    & $0.38$ \\
        \bottomrule
    \end{tabular}
    \caption{Overview of augmentation protocols for heavy, medium, and light intensities.}
    \label{tab:sup_augmentations}
\end{table*}

\section{Encoder Self-Attention Analysis}
\begin{figure*}[h!]
    \centering
    \includegraphics[width=0.8\linewidth]{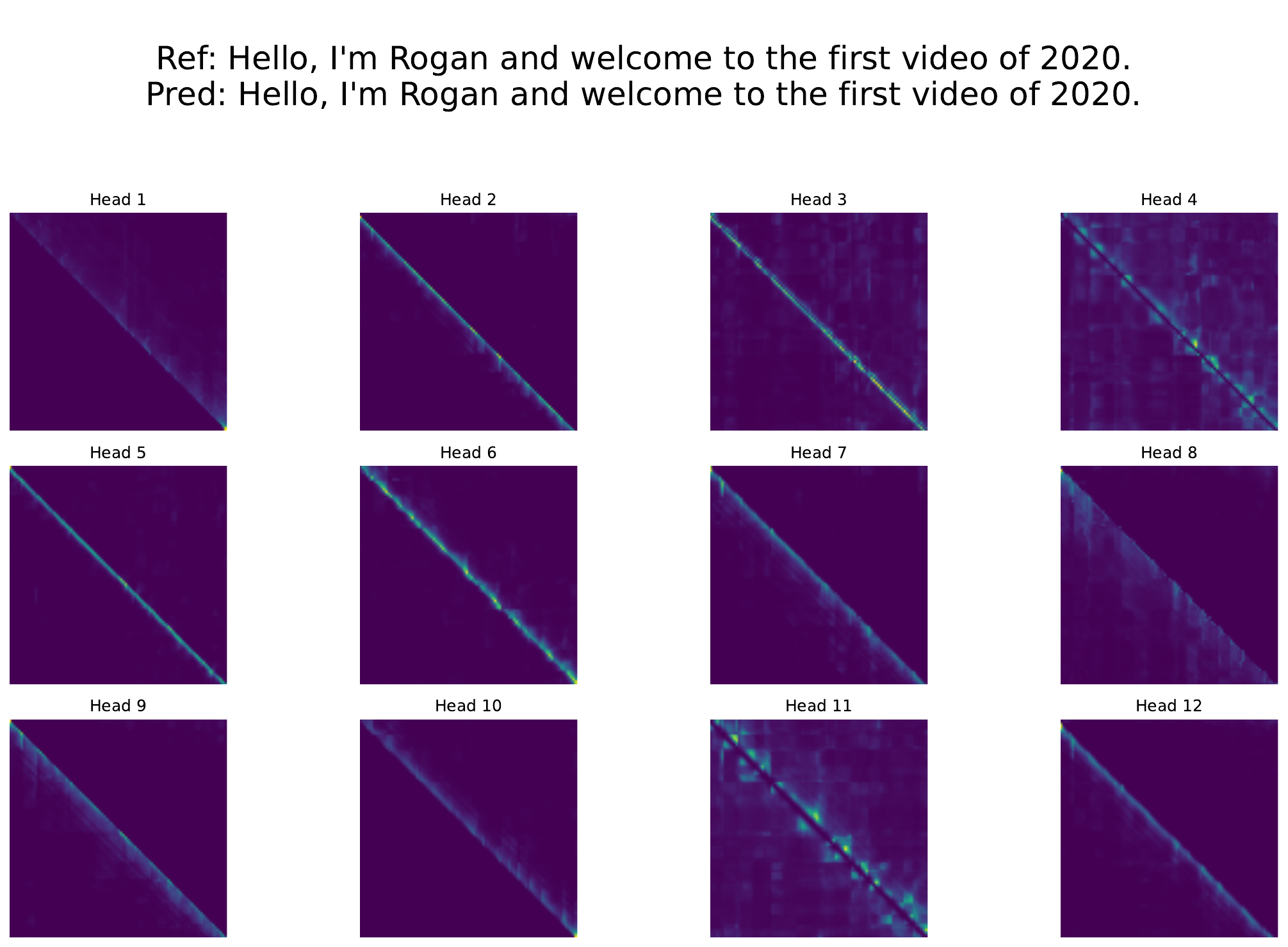}
    \caption{Encoder self-attention averaged over layers per attention head.}
    \label{fig:enc-attn1}
\end{figure*}

\begin{figure*}[]
    \centering
    \includegraphics[width=0.8\linewidth]{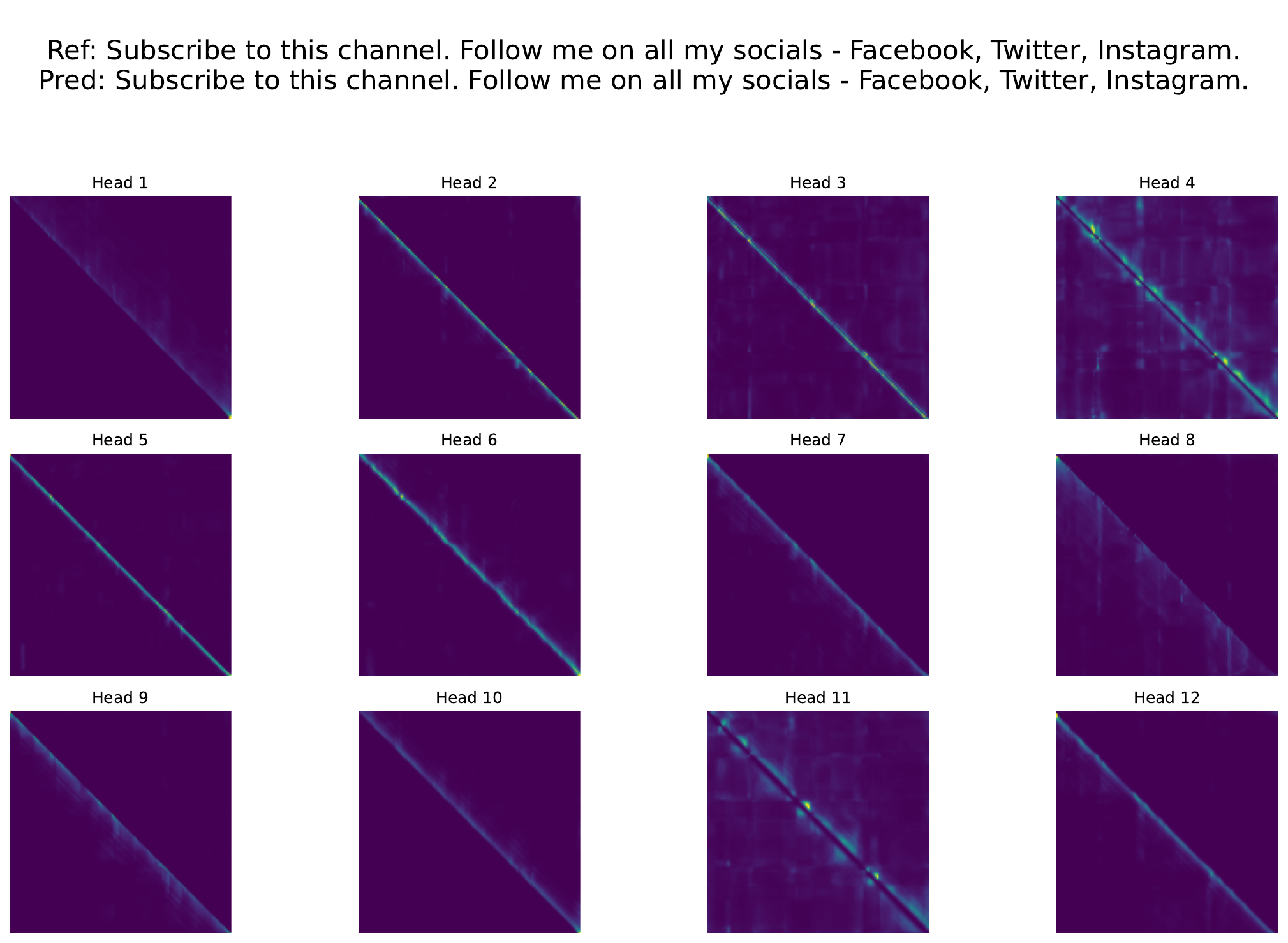}
    \caption{Encoder self-attention averaged over layers per attention head.}
    \label{fig:enc-attn2}
\end{figure*}

\begin{figure*}[]
    \centering
    \includegraphics[width=0.8\linewidth]{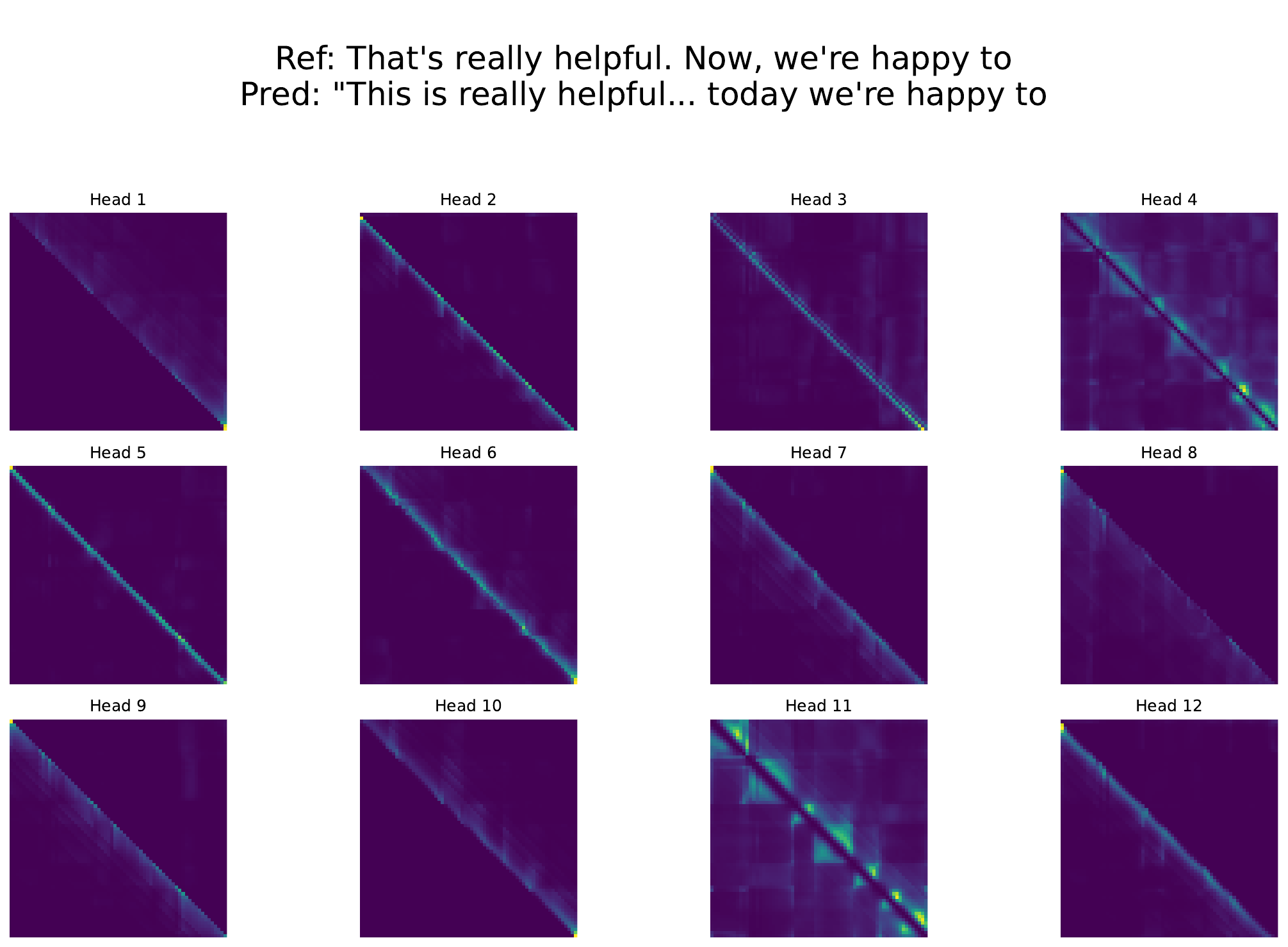}
    \caption{Encoder self-attention averaged over layers per attention head.}
    \label{fig:enc-attn3}
\end{figure*}

Additional examples of head specialization patterns in the encoder attention mechanism during T5 inference are visualized in Figures \ref{fig:enc-attn1}, \ref{fig:enc-attn2} and \ref{fig:enc-attn3}.

\section{Cross-Attention Behavior Phenomena}
\begin{figure*}[]
    \centering
    \includegraphics[width=1\linewidth]{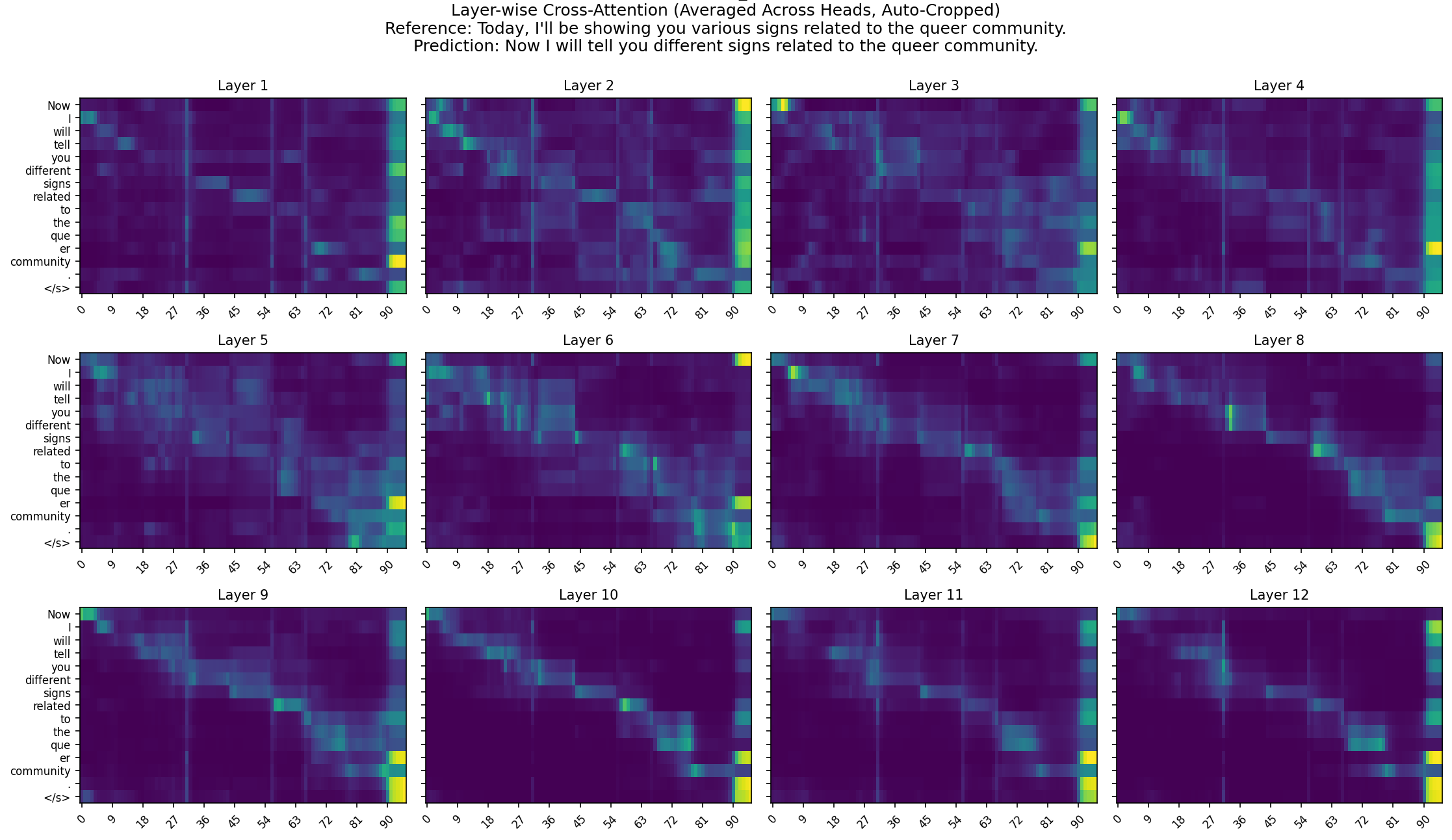}
    \caption{Cross-Attention averaged for each layer over all attention heads, showing temporal progression of tokens attending to frames.}
    \label{fig:cross-attn_layers}
\end{figure*}

\begin{figure*}[]
    \centering
    \includegraphics[width=1\linewidth]{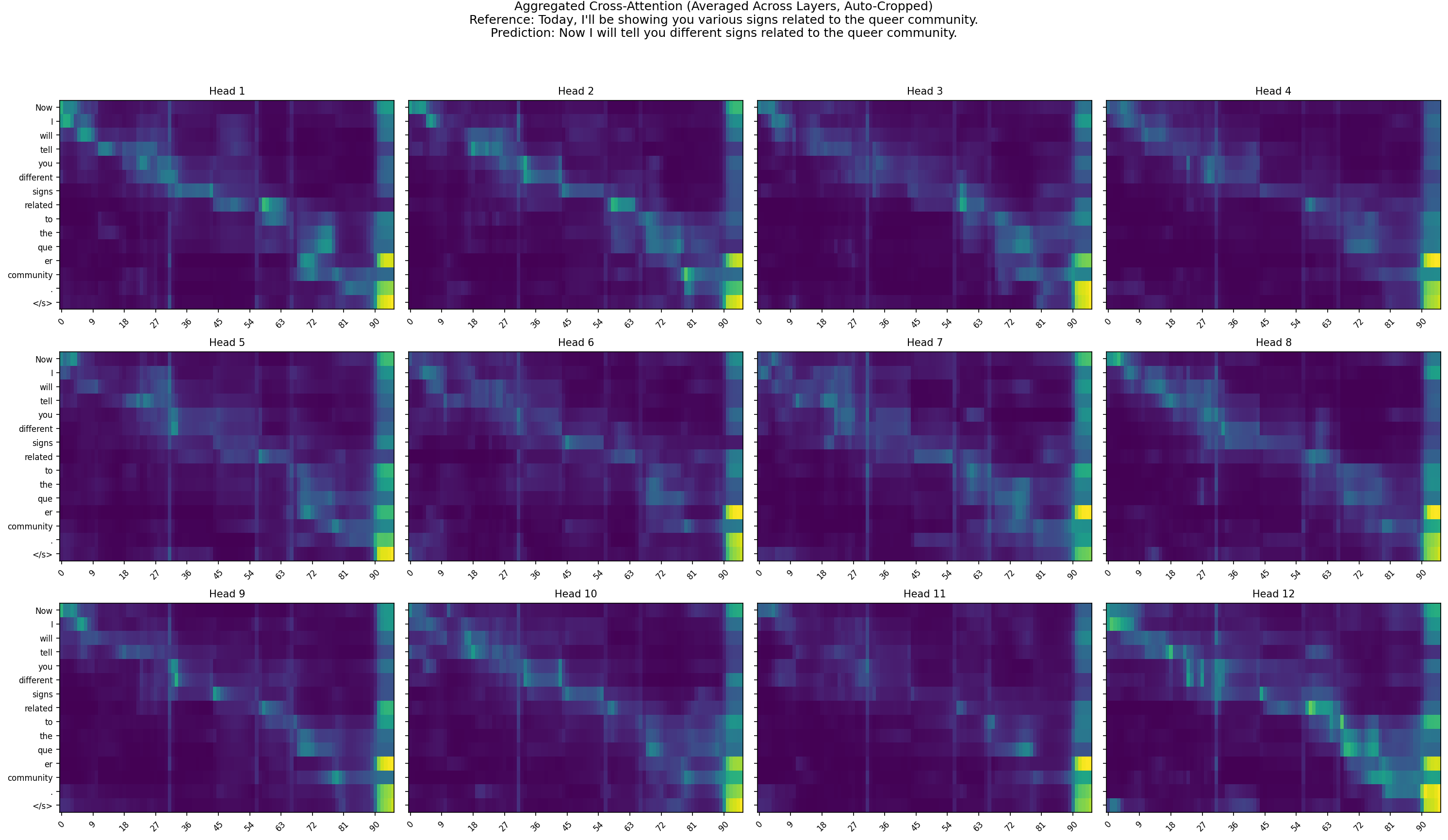}
    \caption{Cross-Attention averaged for each attention head over all layers, showing temporal progression of tokens attending to frames.}
    \label{fig:cross-attn_heads}
\end{figure*}

\begin{figure*}[]
    \centering
    \includegraphics[width=1\linewidth]{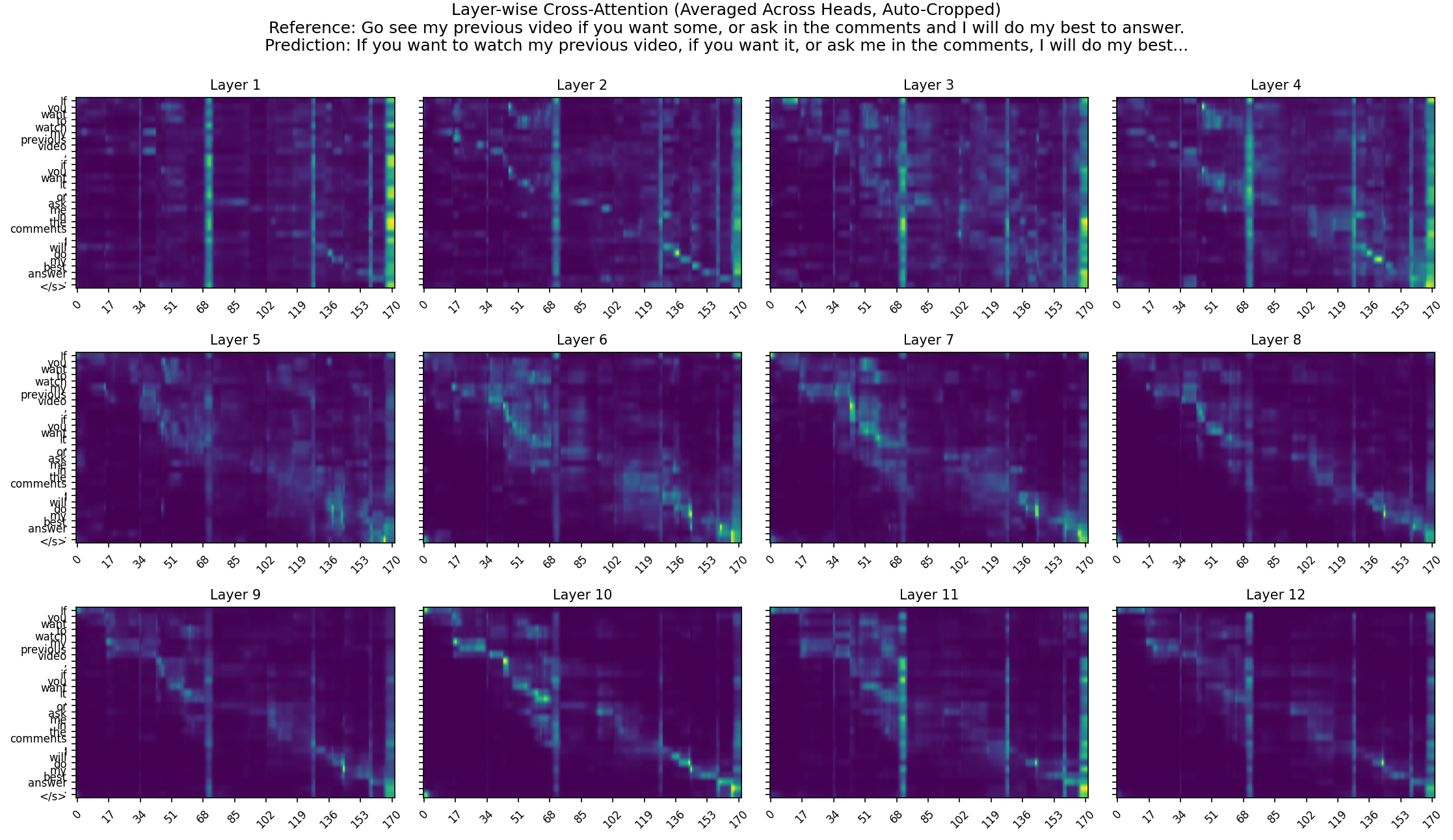}
    \caption{Cross-Attention averaged for each layer over all attention heads, showing temporal progression of tokens attending to frames.}
    \label{fig:cross-attn_layers2}
\end{figure*}

\begin{figure*}[]
    \centering
    \includegraphics[width=1\linewidth]{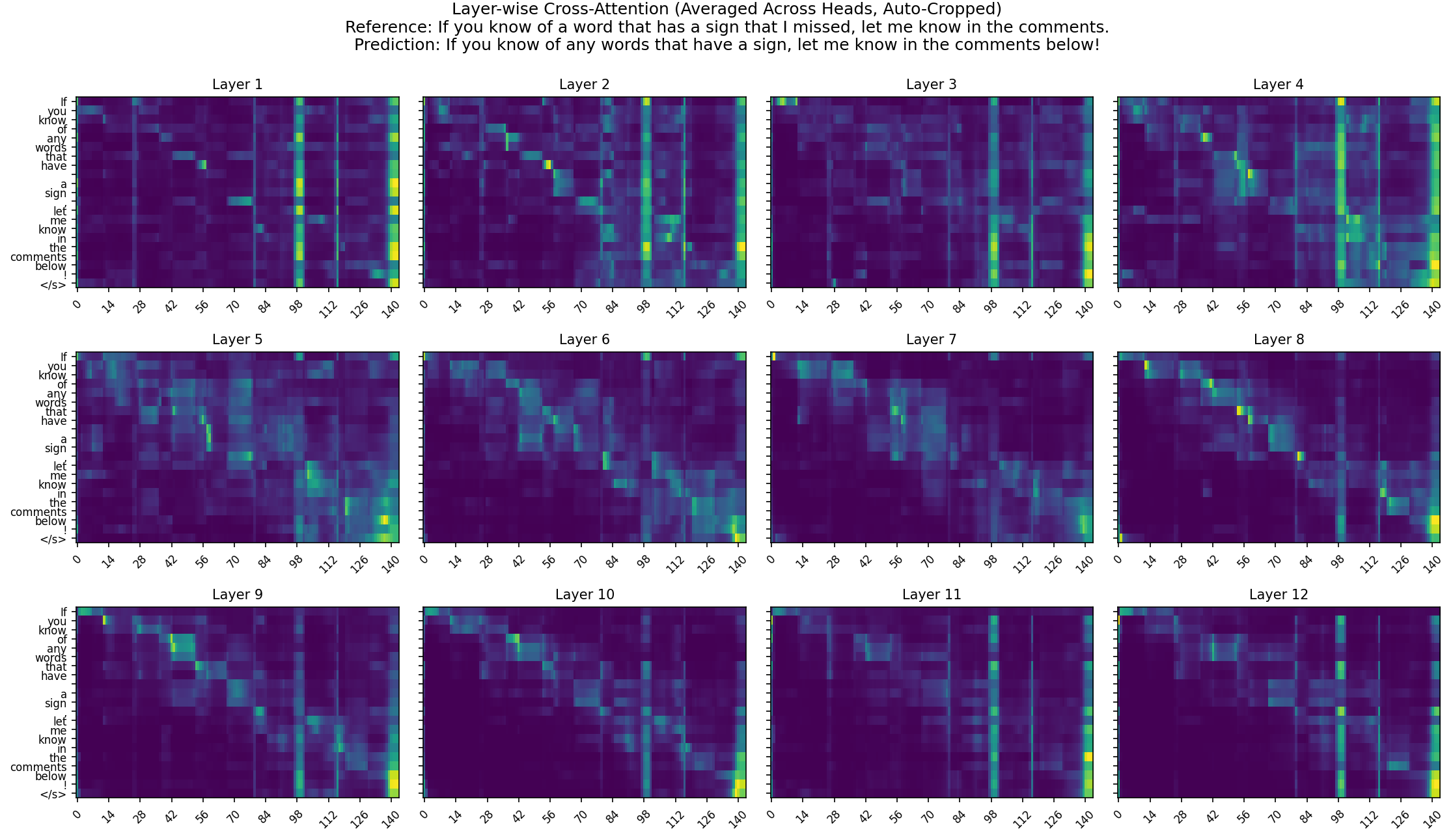}
    \caption{Cross-Attention averaged for each layer over all attention heads, showing temporal progression of tokens attending to frames.}
    \label{fig:cross-attn_layers3}
\end{figure*}

The full visualization of all layer-average cross-attention matrices from inference of a clip translation is shown in Fig. \ref{fig:cross-attn_layers}. In Figure \ref{fig:cross-attn_heads}, cross-attention matrices are averaged for each attention head over the layers, showing the same pattern. Two additional examples are provided in Figures \ref{fig:cross-attn_layers2} and \ref{fig:cross-attn_layers3}.

\begin{figure*}[]
    \centering
    \begin{subfigure}{\linewidth}
        \centering
        \includegraphics[width=\linewidth]{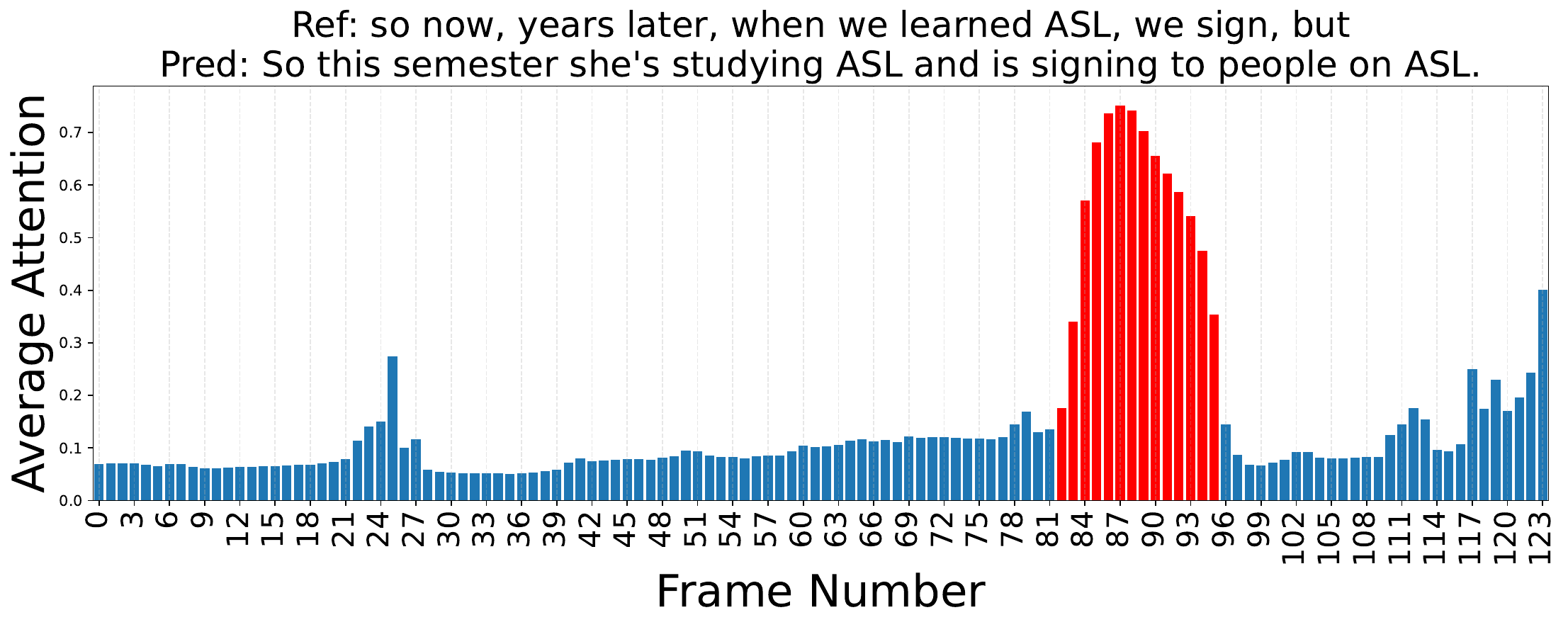}
        \caption{}
        \label{fig:cross-attn_sequence_histogram}
    \end{subfigure}
    
    \begin{subfigure}{\linewidth}
        \centering
        \includegraphics[width=\linewidth]{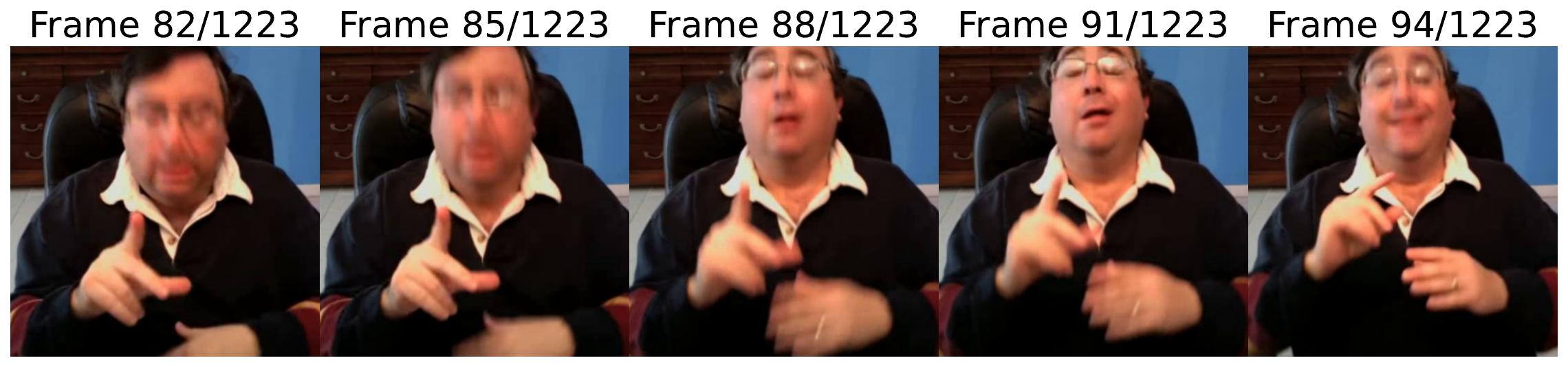}
        \caption{}
        \label{fig:cross-attn_sequence_frames}
    \end{subfigure}
    
    \caption{Histogram (a) visualizes Cross-Attention Distribution over all attention heads and layers, with a long intensity spike sequence in frames 82-95, highlighted in red. Video frames (b) show this is a sequence of mostly still, non-informative frames where the signer didn't change his pose.}
    \label{fig:cross-attn_sequence}
\end{figure*}

In Figure \ref{fig:cross-attn_sequence} we show an additional example of the analyzed behavior where the T5 model is, according to our hypothesis, using non-informative frame segments to encode information about the translation.

